\RequirePackage{amsmath}
\documentclass[runningheads]{llncs}
\usepackage[T1]{fontenc}
\usepackage{graphicx}

\usepackage%[hidelinks]
    {hyperref}
\usepackage{color}

\usepackage{amssymb}
\usepackage{amsfonts}
\usepackage{xspace}
\usepackage[normalem]{ulem}
\usepackage[noend,ruled,linesnumbered]{algorithm2e}
\usepackage{placeins}
\SetKwInOut{Input}{input}
\let\oldnl\nl
\newcommand{\nlnonumber}{\renewcommand{\nl}{\let\nl\oldnl}}

\usepackage{array}
\usepackage{booktabs}
\usepackage{makecell}
\usepackage{multirow}
\usepackage{tikz}
\usetikzlibrary{arrows.meta, shapes, positioning}

\usepackage{float}

\usepackage{pgfplots}
\pgfplotsset{compat=1.18}

\newcommand{\A}{\mathcal{A}}
\newcommand{\C}{\mathcal{C}}

\newcommand{\E}{\mathcal{E}}
\newcommand{\F}{\mathcal{F}}
\newcommand{\G}{\mathcal{G}}

\renewcommand{\L}{\mathcal{L}}

\renewcommand{\P}{\mathcal{P}}
\newcommand{\R}{\mathcal{R}}

\newcommand{\T}{\mathcal{T}}
\newcommand{\Z}{\mathcal{Z}}

\newcommand{\alphabet}{\mathrm{\Sigma}}
\newcommand{\conceptSet}{\alphabet_C}
\newcommand{\roleSet}{\alphabet_R}
\newcommand{\constSet}{\alphabet_I}
\newcommand{\varSet}{\alphabet_V}
\newcommand{\ISA}{\sqsubseteq}

\newcommand{\aczero}{\mathrm{AC}^0}

\newcommand{\coNP}{\textrm{coNP}}
\newcommand{\NL}{\textrm{NL}}

\newcommand{\OWLTWO}{\text{OWL~2}\xspace}

\newcommand{\OWLQL}{\text{OWL~2~QL}\xspace}

\newcommand{\decProb}{\textsc{IGA-Ent}\xspace}

\newcommand{\tup}[1]{\langle #1 \rangle}
\newcommand{\vseq}[1]{\mathbf{#1}}

\newcommand{\ra}{\rightarrow}
\newcommand{\vars}{\textit{vars}}
\newcommand{\atoms}{\textit{QA}}
\newcommand{\pred}{\textit{pred}}

\newcommand{\atomrewr}{\mathsf{AtomRewr}}

\newcommand{\pr}{\Pi}

\newcommand{\wrt}{w.r.t.\xspace}
\newcommand{\cnj}{\gamma}

\newcommand{\modelsiga}{\models_\mathsf{IGA}}
\newcommand{\optcens}{\mathsf{optCens}}

\newcommand{\dlliter}{\text{DL-Lite}_\R}

\newcommand{\allatoms}{Atoms}
\newcommand{\clashformula}{\mathsf{Clash}}
\newcommand{\igaformula}{\mathsf{IGA\text{-}Ent}}

\newcommand{\TGD}{\mathit{TGD}}

\newcommand{\policyexp}{\mathsf{PolicyExp}}
\newcommand{\eq}{\mathsf{eq}}
\newcommand{\conj}{\mathsf{conj}}

\newcommand{\map}{\mathsf{map}}
\newcommand{\body}{\mathsf{body}}
\newcommand{\head}{\mathsf{head}}
\newcommand{\ucqrew}{\mathsf{UCQRew}}

\newcommand{\isdisclformula}{\mathsf{isDiscl}}

\newcommand{\censiga}{\C_{\mathsf{IGA}}}

\newcommand{\clt}[1]{\mathsf{cl}_{\T}(#1)}
\newcommand{\modelseql}{\models_{\mathsf{EQL}}}

\newcommand{\qedex}{\hfill\ensuremath{\vrule height 4pt width 4pt depth 0pt}}

\newenvironment{proofsk}{\noindent \textsl{Proof (sketch).\ }}{\qed}
\newenvironment{forceproof}{\noindent \textsl{Proof.\ }}{\qed}

\newif\ifdraft
\drafttrue %uncomment for draft mode

\newif\ifshowfullproofs
\newif\ifcameraready
\camerareadytrue

\newif\iflong
\longtrue

\ifdraft
    \usepackage{color}
    \newcommand{\nbwidth}{35mm}
    \marginparwidth=\nbwidth
    \newcommand{\nb}[2][auto]{%
      \textcolor{red}{\bf!}%
      \ifthenelse{\equal{#1}{right}}{%
        \reversemarginpar%
        \marginpar{\parbox{\nbwidth}{\raggedright\scriptsize\textcolor{red}{#2}}}%
      }{%
        \ifthenelse{\equal{#1}{left}}{%
          \normalmarginpar%
          \marginpar{\parbox{\nbwidth}{\raggedleft\scriptsize\textcolor{red}{#2}}}%
        }{%
          \marginpar[\parbox{\nbwidth}{\raggedleft\scriptsize\textcolor{red}{#2}}]%
        	{\parbox{\nbwidth}{\raggedright\scriptsize\textcolor{red}{#2}}}%
        }%
      }%
    }
    \else
    \newcommand{\nb}[1]{}
\fi

\begin{document}
    \title{%
    \iflong
        CQE under Epistemic Dependencies:\\Algorithms and Experiments\\(extended version)
    \else
        Controlled Query Evaluation under Epistemic Dependencies: Algorithms and Experiments
    \fi}
    \titlerunning{CQE under Epistemic Dependencies: Algorithms and Experiments}
    \author{%
        Lorenzo Marconi%\inst{1}
        \orcidID{0000-0001-9633-8476} 
        \and
        Flavia Ricci%\inst{1}
        %\orcidID{????-????-????-????}
        \and
        \newline
        Riccardo Rosati%\inst{1}
        \orcidID{0000-0002-7697-4958}
    }
    \authorrunning{L.\ Marconi et al.}
    \institute{%
        Sapienza University of Rome, Italy\\
        \email{\{marconi,rosati\}@diag.uniroma1.it}\\
        \email{ricci.1883245@studenti.uniroma1.it}
    }
    \maketitle              % typeset the header of the contribution
    \begin{abstract}
        % The abstract should briefly summarize the contents of the paper in 150--250 words.
        We investigate Controlled Query Evaluation (CQE) over ontologies, where information disclosure is regulated by epistemic dependencies (EDs), a family of logical rules recently proposed for the CQE framework. In particular, we combine %the usage of 
        EDs with the notion of optimal GA censors, i.e.\ maximal sets of ground atoms that are entailed by the ontology and can be safely revealed. We focus on answering Boolean unions of conjunctive queries (BUCQs) with respect to the intersection of all optimal GA censors---an approach that has been shown in other contexts to ensure strong security guarantees with favorable computational behavior.
        First, we characterize the security of this intersection-based approach and identify a class of EDs (namely, full EDs) for which it remains safe. Then, for a subclass of EDs and for $\dlliter$ ontologies, we show that answering BUCQs in the above CQE semantics is in $\aczero$ in data complexity by presenting a suitable, detailed first-order rewriting algorithm.
        Finally, we report on experiments conducted in two different evaluation scenarios, showing the practical feasibility of our rewriting function.
        
    \end{abstract}

    \section{Introduction}
\label{sec:introduction}

%\nb{Lor: Tutti i minor fix sono stati fatti. Rivedere cosa abbiamo promesso nei \href{https://docs.google.com/document/d/1OCwP3b7BswfXkyY9x64r_MpH35U_-HjhK88LCdaljzs}{rebuttal} e implementare il più possibile (ora abbiamo 16 pagine).}
The ever-growing volume of structured and semantically rich data has created new challenges for knowledge management and data security. Several modern applications in domains like healthcare and finance rely on ontologies to establish shared vocabularies and formal semantics, facilitating effective data organization and retrieval. While these tools provide sophisticated querying and inference, they also raise critical concerns about \iflong information disclosure: sensitive facts may be unintentionally revealed through apparently harmless queries when the underlying ontological axioms are taken into account.
\else unintended information disclosure: apparently harmless queries can reveal sensitive data when combined with ontological knowledge. \fi
\emph{Controlled Query Evaluation} (CQE)~\cite{Bisk00,BiBo04,BoSa13,CKKZ13} is a framework that addresses these concerns by mediating access to data in such a way that only information compliant with a formal \emph{data protection policy}---expressed in logical terms---is accessible through queries.

\iflong This work applies \else We apply\fi CQE to ontologies based on Description Logics (DLs)\footnote{For an up-to-date overview of CQE in the context of DLs, we refer the reader to~\cite{Bona22,CLMRS24-SNCS}.}~\cite{BCMNP07},
\iflong
a family of logics, many of which expressible in first-order (FO) logic, for which the most important reasoning problems are usually decidable. In DLs, 
\else
where 
\fi
knowledge is structured into a TBox, containing intensional axioms, and an ABox, containing extensional facts.
The pivotal notion in CQE is the one of \emph{censor}, a set of logical formulas that are logically implied by the ontology and %, at the same time, 
comply with the policy. Specifically, we are interested in \emph{GA censors}~\cite{CLRS24}, which consist of ground atoms, hence structurally resembling ABoxes or relational databases.

Usually, CQE policies are defined using \emph{denials}, i.e.\ expressions of the form $(\exists \vseq{x}\,\gamma(\vseq{x}))\rightarrow\bot$, where $\exists \vseq{x}\,\gamma(\vseq{x})$ is a \emph{Boolean conjunctive query} (BCQ). Denials are used to specify information that must remain undisclosed: the system must prevent users from inferring that the formula $\exists \vseq{x}\,\gamma(\vseq{x})$ holds in the ontology.
\ifcameraready The recent work~\cite{CLMRS24} \else In the recent work~\cite{CLMRS24}, the authors \fi proposed an extension for such a language of rules, called \emph{epistemic dependencies} (EDs)~\cite{CL20}, which are logical implications between two (possibly open) conjunctive queries, each within the scope of a modal operator $K$, though adopting a notion of censor that differs from the one considered in the current paper.

\newcommand{\predSymbol}[1]{\allowbreak\textsf{#1}}
\newcommand{\salary}{\predSymbol{salary}}
\newcommand{\manager}{\predSymbol{manager}}
\newcommand{\managerOf}{\predSymbol{managerOf}}
\newcommand{\managesDept}{\predSymbol{respDept}}
\newcommand{\consRel}{\predSymbol{consRel}}
\newcommand{\indA}{\predSymbol{lucy}}
\newcommand{\indB}{\predSymbol{tom}}
\newcommand{\salaryA}{\predSymbol{150k}}
\newcommand{\salaryB}{\predSymbol{75k}}
\begin{example}
\label{ex:init}
    The policy of a company stipulates that all salaries of employees must be kept confidential, except those of managers.
    In addition, the existence of consensual personal relationships between managers and their employees must remain undisclosed.
    
    In logical terms, such a policy can be defined as the following set of EDs:
    %\[
    %\begin{array}{r@{}l}
         % \P=\{\, \forall x,y\,(&K \salary(x,y)\rightarrow K\manager(x)),\\
         % \forall x,y\,(&K \managerOf(x,y)\land\consRel(x,y)\rightarrow K \bot) \,\}
    %\end{array}
    %\]
    \[
    \begin{array}{r@{\,}l}
         \P = \{&\forall x,y\,(K \salary(x,y) \rightarrow K\manager(x)),\\
         & K \exists x,y\,(\managerOf(x,y)\land\consRel(x,y)) \rightarrow K \bot \,\}
    \end{array}
    \]
    \iflong
    where $\manager$ is a unary predicate indicating that an individual is a manager, and $\salary$, $\consRel$ and $\managerOf$ are binary predicates modelling, respectively, the salary level of a person, the consensual relationship between two individuals and the relationship where one individual manages another.
    \else
    where $\manager$ is a unary predicate and $\salary$, $\consRel$ and $\managerOf$ are binary predicates with the intuitive meaning.
    \fi
    In the second ED, the usage of the existential quantifier indicates that, for every manager (resp., employee), the very existence of a consensual relationship with any employee (resp., manager) of hers must not be revealed---not merely the identities of the individual involved.

    Moreover, suppose that the company ontology consists of:
    \begin{itemize}
        \item A TBox $\T=\{\exists\managerOf\ISA\manager,\manager\ISA\exists\managesDept\}$, meaning that everyone who manages another individual is a manager, and managers are such only if they are responsible for some department.
        \item An ABox $\A=\{\managerOf(\indA,\indB),\consRel(\indA,\indB),\salary(\indA,\salaryA),\salary(\indB,\salaryB)\}$, meaning that Lucy is Tom's manager, they have a consensual relationship, and their salary is \$150,000 and \$75,000, respectively.
    \end{itemize}
    A censor consisting only of ground atoms must remove at least one of the facts $\managerOf(\indA,\indB),\consRel(\indA,\indB)$ from $\A$ and, at the same time, must remove the fact $\salary(\indB,\salaryB)$ (because Tom is not a manager).
    By contrast, any optimal GA censor can safely include the facts $\manager(\indA)$ and $\salary(\indA,\salaryA)$, because Lucy’s managerial status follows from the ontology and knowing that she is a manager (and her salary) does not violate the policy.
    \qedex
\end{example}

Our main objective is to evaluate queries under a formal entailment semantics that maximizes data disclosure while remaining compliant with the policy. Thus, we call \emph{optimal GA censors} the GA censors that are maximal w.r.t.\ set inclusion, and focus on the problem of checking whether a \emph{Boolean union of conjunctive queries} (BUCQ) is entailed by the TBox and the intersection of all the optimal GA censors.
%%
%From the results of~\cite{CLRS20} it follows that this task, known as IGA-entailment,
This task, known as \emph{IGA-entailment}, has been shown to be \emph{FO-rewritable} when the TBox is expressed in $\dlliter$ and the policy consists of denials~\cite{CLMRS25-JoWS}.
That is, in the above setting, IGA-entailment of a BUCQ $q$ can be decided by rewriting $q$ into a new FO query $q_r$ that only depends on the TBox and the policy and, in a second moment, evaluating $q_r$ over the ABox.
This property guarantees a nice computational behaviour at a theoretical level, as the task enjoying it has the same complexity as evaluating an SQL query over a database. However, it still needs to be empirically validated through a practical implementation. % that demonstrates the efficiency of the rewriting system used to generate $q_r$.
In the case based on denials, a working prototype was provided in~\cite{CLMRS20}, within the ontology-based data access framework.

We aim to extend this scenario to accommodate policies defined using EDs while preserving the \emph{FO-rewritability} property.
First, we prove that the class of \emph{full EDs} and \emph{linear EDs} enjoy a desirable property related to security. We exclude, however, the possibility that IGA-entailment remains FO-rewritable for such classes of dependencies, by proving \coNP- and \NL-hardness results for the related decision problem, respectively.
We thus identify a condition for full EDs for which we are able to prove the FO-rewritability.
For two classes of EDs that respect such a condition, namely the linear full and the acyclic full EDs, we finally conducted experiments to test the practical feasibility of our rewriting algorithm. Specifically, we implemented a tool that rewrites a SPARQL BUCQ into a new query $q_r$ solely based on the given TBox and policy, and then evaluates $q_r$ over an SQL database containing the ABox.
Since our theoretical results are related to the logic $\dlliter$, we adopted the \OWLQL ontology of the OWL2Bench benchmark~\cite{SBM20} as our testbed.
Two distinct evaluation scenarios demonstrate that our method is not only theoretically sound but also practically feasible, with most rewritten queries running within acceptable time bounds.

% This paper provides the following key contributions to the CQE framework:
% \begin{itemize}
%     \item We explored the use of epistemic dependencies (EDs) for the CQE framework, delineating new theoretical results including a computational complexity analysis over different classes of EDs.
%     \item We identified a class of EDs for which it is possible to reduce the problem studied to the evaluation of a first-order query over an ABox/database.
%     \item We implemented our first-order rewriting algorithm for enabling policy-protected query answering over large ontologies and tested it over a benchmark for OWL ontologies.
% \end{itemize}

The paper is structured as follows.
Section~\ref{sec:preliminaries} provides the necessary theoretical background. Section~\ref{sec:framework} describes the framework and the problem studied.
Section~\ref{sec:security-of-intersection} introduces a key property that helps identify interesting subclasses of EDs.
Section~\ref{sec:negative-results} presents lower bounds that exclude FO-rewritability in the general case.
Section~\ref{sec:fo-rewritability} defines the subclass of EDs we target and presents a detailed FO-rewriting algorithm.
Section~\ref{sec:experiments} reports our experimental findings.
Finally, Section~\ref{sec:conclusions} concludes the paper.
    \section{Preliminaries}
\label{sec:preliminaries}

In this paper, we refer to standard notions of function-free first-order (FO) logic and Description Logics (DL). We use countably infinite sets of symbols $\conceptSet$, $\roleSet$, $\constSet$ and $\varSet$, containing respectively unary predicates (called \emph{concepts}), binary predicates (called \emph{roles}), constant symbols (also called \emph{individuals}) and variables.
An \emph{atom} is a formula of the form $P(\vseq{t})$, where $P$ is a (either unary or binary) predicate and $\vseq{t}$ is a sequence of terms, i.e.\ variables or constants.

Given a set of FO formulas $\Phi$, we denote by $\vars(\Phi)$ and $\pred(\Phi)$, the sets of variables and predicates occurring in $\Phi$, respectively.
Given any FO formula $\phi$, we use the notation $\phi(\vseq{x})$ when we want to emphasize its free variables $\vseq{x}$, and we overload $\vars(\cdot)$ to work with FO formulas, with the same meaning.
If $\phi(\vseq{x})$ is closed (that is, $\vseq{x}$ is empty), then it is called a \emph{sentence}; Furthermore, if $\vars(\phi)=\emptyset$, then $\phi$ is said to be \emph{ground}. In particular, ground atoms are also called \emph{facts}.
If $\F$ is a set of facts, we say that an FO sentence $\phi$ evaluates to true in $\F$ to actually mean that it is true in the Herbrand model of $\F$.

%In this work we also refer to specific classes of domain-independent FO formulas, such as the one of \emph{conjunctive queries} (CQs), i.e.\ conjunctions of atoms possibly occurring in the scope of an existential quantifier.
In this work, we also refer to specific classes of domain-independent FO formulas, such as the class of \emph{conjunctive queries} (CQs), i.e.\ formulas of the form $\exists\vseq{x}\,(\gamma)$, where $\vseq{x}\subseteq\vars(\gamma)$ and $\gamma$ is a conjunction of atoms.
A disjunction of CQs sharing the same free variables is called \emph{union of conjunctive queries} (UCQ), which sometimes we also treat as a set of CQs.
As customary, closed CQs and UCQs are said to be \emph{Boolean} and referred to as BCQs and BUCQs, respectively.
Given any CQ $q$, we indicate by $\atoms(q)$ the set of atoms of $q$.

We call \emph{substitution} any function $\sigma:\varSet\to\varSet\cup\constSet$. Given a CQ $q$ and a substitution $\sigma$ of (a subset of) its variables, we write $\sigma(q)$ to denote the result of applying $\sigma$ to $q$. 
A substitution of variables is said to be \emph{ground} if its image is contained in $\constSet$.
Moreover, given a set of facts $\F$, an \emph{instantiation} for $q$ in $\F$ is a ground substitution $\sigma$ of the variables of $q$ such that $\atoms(\sigma(q))\subseteq\F$. If such an instantiation $\sigma$ exists, the set $\atoms(\sigma(q))$ is called \emph{image} of $q$ in $\F$.
Given a UCQ $q$, an \emph{image of $q$ in $\F$} is any image of $q'$ in $\F$, for any $q'\in q$.

We resort to DL ontologies as a formal way of representing structured knowledge about a given domain. Ontologies are usually partitioned into two sets %, called TBox and ABox, 
used, respectively, for representing intensional and extensional knowledge.
More formally, given a DL $\L_\T$, an $\L_\T$ \emph{ontology} is a finite set $\T\cup\A$, where $\T$ (called $\L_\T$ TBox) is a set of axioms expressible in $\L_\T$, and $\A$ (called ABox) is a set of ground atoms.
In particular, our complexity results hold for ontologies expressed in $\dlliter$~\cite{CDLLR07}, which is the logic underpinning \OWLQL, one of the three \OWLTWO profiles~\cite{CHMP*08,W3Crec-OWL2-Profiles} that is specifically designed for efficient query answering.
The axioms of a $\dlliter$ TBox $\T$ take the 
\iflong
following form:
\[B \ISA B', \quad R \ISA R', \quad B \ISA\neg B', \quad R \ISA\neg R'\]
\else
form $B \ISA B'$, $R \ISA R'$, $B \ISA\neg B'$, or $R \ISA\neg R'$,
\fi
where $B$ and $B'$ (resp., $R$ and $R'$) are of the form $A$, $\exists S$ or $\exists S^-$ (resp., of the form $S$ or $S^-$), with $A\in\conceptSet$, $S\in\roleSet$ and $S^-$ the inverse of $S$. The unqualified existential restriction $\exists S$ (resp., $\exists S^-$) represents the set of individuals occurring as the first (resp., second) argument of $S$.
%%
%It is well known~\cite{CDLLR07} that standard BUCQ entailment under $\dlliter$ ontologies is FO-rewritable, i.e., given a $\dlliter$ TBox $\T$ and a BUCQ $q$, there exists an FO query (actually, a BUCQ) $q_r$ such that, for every ABox $\A$, $\T \cup \A \models q$ holds iff $q_r$ evaluates to true in $\A$.
%In the following, we use a function $\ucqrew(q,\T)$ to indicate such $q_r$.

% In this work, we make use of the algorithm $\perfectref$, for which the following property is well-known to hold:
% \begin{proposition}[\cite{CDLLR07}]
% \label{pro:dllite-qa}
%     Let $\T$ be a $\dlliter$ TBox and $q$ be a BUCQ. For every ABox $\A$, we have that $\T \cup \A \models q$ if and only if $\perfectref(q,\T)$ evaluates to true in $\A$.
% \end{proposition}

Moreover, we denote with $\clt{\A}$ the \emph{closure} of $\A$ w.r.t.\ $\T$, i.e.\ the set of all the ground atoms that are logical consequences of $\T\cup\A$.
We also refer to the rewriting function $\atomrewr$, for which the following property has been demonstrated in~\cite[Lemma~6]{CLRS24}.

\begin{proposition}[\cite{CLRS24}]
\label{pro:atomrewr}
    Let $\T\cup\A$ be a consistent $\dlliter$ ontology, and let $\phi$ be an FO sentence. Then, $\phi$ evaluates to true in $\clt{\A}$ if and only if $\atomrewr(\phi,\T)$ evaluates to true in $\A$.
\end{proposition}

Finally, Section~\ref{sec:negative-results} requires the reader to be familiar with basic notions of computational complexity theory~\cite{Papa94}. Indeed, decision problems solvable through FO-rewriting are known to be in $\aczero$ in data complexity~\cite{AbHV95}, and we refer to the complexity classes $\NL$ and $\coNP$ (both of which are known to be strict supersets of $\aczero$) for showing scenarios in which the studied problem is not FO-rewritable.
    \section{Framework}
\label{sec:framework}

In the spirit of~\cite{CLMRS24}, we adopt as \emph{protection policy} (or simply \emph{policy)} a finite set of epistemic dependencies, which are a special case of EQL-Lite(CQ)~\cite{CDLLR07b} sentences, and are defined as follows.

\begin{definition}
\label{def:epistemic-dependency}
    An \emph{epistemic dependency (ED)} is a sentence $\tau$ of the \iflong following\fi form:
    \begin{equation}
    \label{eqn:epistemic-dependency}
        \begin{array}{l}
        \forall \vseq{x}_1,\vseq{x}_2\,(K q_b(\vseq{x}_1,\vseq{x}_2) \ra K q_h(\vseq{x}_2))
        \end{array}
    \end{equation}
    where $q_b(\vseq{x}_1, \vseq{x}_2)$ is a CQ with free variables $\vseq{x}_1 \cup \vseq{x}_2$, $q_h(\vseq{x}_2)$ is a CQ with free variables $\vseq{x}_2$%
    % \iflong
    , and $K$ is a modal operator. %The variables $\vseq{x}_2$ are called the frontier variables of $\tau$.
    % \else
    % \ $K$ is a modal operator, and $\vseq{x}_2$ are called the frontier variables of $\tau$.
    % \fi
    Given an ED $\tau$, we refer to the CQs $q_b$ and $q_h$ with $\body(\tau)$ and $\head(\tau)$, respectively.
\end{definition}
We say that an FO theory $\Phi$ \emph{satisfies} an ED $\tau$ (in symbols $\Phi\modelseql\tau$) if, for every ground substitution $\sigma$ for the free variables of $\body(\tau)$, if $\Phi\models\sigma(\body(\tau))$ then $\Phi\models\sigma(\head(\tau))$. If all the EDs of a policy $\P$ are satisfied by $\Phi$, then we say that $\Phi$ \emph{satisfies} $\P$  (in symbols $\Phi\modelseql\P$).
Intuitively, EDs express the disclosure rules that should govern the publication of data. For every ground substitution $\sigma$ of the universal variables of an ED $\tau$, if the ontology entails $\sigma(\body(\tau))$, then this information may only be disclosed if $\sigma(\head(\tau))$ can also be disclosed. In the special case in which $\head(\tau)=\bot$, the ED acts as a denial constraint, i.e.\ its body must not be revealed under any instantiation of its free variables. 

The input to our framework is a triple $\tup{\T,\P,\A}$, called \emph{CQE instance} (or \emph{$\L$ CQE instance}, if $\T\cup\A$ is expressed in a given DL $\L$), where $\T$ is a DL TBox, $\P$ is a policy and $\A$ is an ABox such that $\T\cup\A$ is consistent.
Given a CQE instance $\tup{\T,\P,\A}$, it may be the case that $\T\cup\A\not\modelseql\P$, i.e.\ the policy is not satisfied by the ontology $\T\cup\A$. For hiding the part of information that should be protected, we rely on the following notion of GA censor~\cite{CLRS24}.
\begin{definition}[GA censor]
\label{def:ga-censor}
    Given a CQE instance $\E=\tup{\T,\P,\A}$, a \iflong \emph{ground atom censor} (in short, \emph{GA censor}) \else \emph{ground atom (GA) censor} \fi for $\E$ is any subset $\C$ of $\clt{\A}$ such that $\T\cup\C\modelseql\P$.
    A GA censor $\C$ of $\E$ is \emph{optimal} if no GA censor $\C'$ of $\E$ exists such that $\C'\supset\C$.
\end{definition}
Informally, an optimal GA censor is a maximal (w.r.t.\ set inclusion) portion of $\clt{\A}$ that can be safely disclosed without violating the given policy.
Among the possibly many GA censors for a CQE instance, we are particularly interested in the information that is common to all optimal GA censors, which represents the maximal amount of facts that can be safely revealed in every scenario where policy satisfaction is preserved. Based on this idea, we now introduce the central notion of IGA-entailment, which formalizes query answering under the intersection of all optimal GA censors.

\begin{definition}[IGA-entailment]
\label{def:iga-entailment}
    Given a CQE instance $\E=\tup{\T,\P,\A}$ and a BUCQ $q$, we say that $\E$ \emph{IGA-entails} $q$ (in symbols $\E\modelsiga q$) if $\T\cup\censiga\models q$, where $\censiga$ is the intersection of all the optimal GA censors of $\E$.
\end{definition}

\begin{example}
\label{ex:iga-entailment}
    Consider the CQE instance $\E=\tup{\T,\P,\A}$, where $\T$, $\P$ and $\A$ are as in Example~\ref{ex:init}. In this case, we have only two optimal censors, i.e.:
    \[
    \begin{array}{l}
        \C_1=\{\managerOf(\indA,\indB),\manager(\indA),\salary(\indA,\salaryA)\}, \\
        \C_2=\{\consRel(\indA,\indB),\manager(\indA),\salary(\indA,\salaryA)\} .
    \end{array}
    \]
    Now, we have that a BCQ like $q_1=\exists x,y\,\consRel(x,y)$ is not IGA-entailed by $\E$, because the extension of $\consRel$ is empty in $\censiga=\C_1\cap\C_2$ and no relevant conclusions can be drawn from the intensional axioms.
    Conversely, the BCQ $q_2=\exists x,y\,(\managesDept(x,y)\land\salary(x,\salaryA))$ is IGA-entailed by $\E$, because the facts $\manager(\indA)$ and $\salary(\indA,\salaryA)$ belong to the intersection $\censiga$.
    \qedex
\end{example}

It is worth noting that an alternative, more expressive kind of entailment has been proposed in the literature, i.e., the skeptical entailment over all the optimal GA censors. However, such an approach has already been shown to be computationally intractable (specifically, \coNP-hard in data complexity) even for BCQs and for policies restricted to sets of denial constraints~\cite[Thm.~6]{CLRS24}.

Although the above definitions apply to any CQE instance, our complexity analysis focuses on specific classes of EDs, which are defined below.
First, borrowing the terminology from the literature on both databases and existential rules (see e.g.\ \cite{BLMS11,CGL12}), an ED $\tau$ is called \emph{full} if no existential variable occurs in its head, and is called \emph{linear} if $|\atoms(\body(\tau))|=1$. A policy is \emph{full} (resp., \emph{linear}) if all its EDs are full (resp., linear).
Then, following the paper~\cite{CLMRS24} that introduced EDs for CQE, we define the notion of acyclicity as follows. Given a policy $\P$ and a TBox $\T$, consider the graph $G$ whose nodes are the predicates of $\T\cup\P$, and whose edges are of two kinds: T-edges, which connect two nodes $A$ and $B$ of $G$ if they occur, respectively, on the left- and right-hand side of a concept or role inclusion of $\T$; analogously P-edges connect two nodes $A$ and $B$ of $G$ if they occur, respectively, in the body and head of an ED of $\P$.
Then, $\P$ is said to be \emph{acyclic for $\T$} if $G$ contains no cycle involving a P-edge.

%We denote by $\classA$, $\classF$ and $\classL$ the classes of acyclic (w.r.t.\ the coupled TBox), full and linear dependencies. 

%In the following, we study the data complexity of the problem of deciding whether $\E\modelsiga q$, given a BUCQ $q$ and a $\dlliter$ CQE instance $\E$.

%We finally define $\decProb$, i.e.\ the decision problem of checking whether a BUCQ is entailed under the IGA semantics.

\iflong
We finally define $\decProb$ as the following decision problem.
{
\noindent
\begin{center}
	\framebox{
		\begin{tabular}{l@{\enspace}l}
			%Problem: & $\decProb[\polClass]$ \\
			%Input: & A $\dlliter$ CQE instance $\E=\tup{\T,\A,\P}$ s.t.\ $\P\in\polClass$, \\ & a BUCQ $q$\\
			Problem: & $\decProb$ \\
			Input: & A $\dlliter$ CQE instance $\E=\tup{\T,\A,\P}$, a BUCQ $q$\\
			Question: & Does $\E\modelsiga q$?
		\end{tabular}
	}
\end{center} 
}
\else
We finally define the decision problem that we focus on, called $\decProb$, as the problem of checking whether a BUCQ is entailed under the IGA semantics.
\fi

    \section{Security of the Intersection of GA Censors}
\label{sec:security-of-intersection}

Before delving into the complexity analysis of the \decProb problem, it is natural to ask whether the intersection of all optimal GA censors constitutes in turn a valid censor, i.e.\ whether it satisfies the given policy.

Actually, the above property does not hold, in general, even in the case when the TBox is empty, as shown by the following example. 

\begin{example}
\label{ex:unsafe-intersection}
    Consider the CQE instance $\E=\tup{\T,\P,\A}$, where $\T$ is empty and
    \[
    \begin{array}{r@{\ }l@{\,}l@{\qquad}r}
        \P =& \{ &
           \tau_1: K(B(1)\land B(2)) \ra K \bot,
           & \A = \{ C(0),B(1),B(2)\,\}.\\
        && \tau_2: \forall x\,(K C(x) \ra K\exists y\,B(y))
        \,\}, \\
        %\A =& \{& C(0),B(1),B(2)\,\}.
    \end{array}
    \]
    We observe that $\C_1=\{C(0),B(1)\}$ and $\C_2=\{C(0),B(2)\}$ are the only two optimal GA censors for $\E$, and so their intersection $\censiga$ is $\{C(0)\}$.
    Then, one can see that $\T\cup\censiga\not\modelseql\tau_2$, which implies that $\censiga$ is not a censor.
    %
    % Now, we first observe that $\T\cup\censiga\not\modelseql\tau_2$, which implies that $\censiga$ is not a censor.
    % At first sight, this might not appear to be a significant issue, as a possible malicious user, by exploiting the knowledge of $\T$, $\P$ and $\censiga$ and by just applying deduction, would only be able to infer that the extension of $B$ is not empty---information that is not intended to be concealed \textit{per se}.
    %
    % However, upon closer examination, the issue turns out to be more serious. Indeed, an attacker can observe that $C(0)$ is exposed because it belongs to all optimal GA censors and, by applying a form of non-standard reasoning, can understand that the only explanation for the absence of any $B$-fact in the intersection is that every optimal GA censor contains either $B(1)$ or $B(2)$ and, at the same time, both such facts belong to at least one censor (and so they are entailed by $\T\cup\A$), thereby concluding a sensitive piece of information (i.e.\ the body of $\tau_1$).
    \qedex
\end{example}

Notice that the policy employed in Example~\ref{ex:unsafe-intersection} is acyclic for the coupled (empty) TBox. However, it is possible to demonstrate that it is safe to refer to the intersection of the optimal GA censors when considering the classes of linear EDs and full EDs, which we focus on in the next section.

\begin{theorem}
    For every CQE instance $\E=\tup{\T,\P,\A}$ such that $\P$ is either full or linear, %it holds that
    % \iflong%
    % \[ \T \cup \bigcap_{\C\in\optcens(\E)}\C \modelseql \P. \]
    % \else%
    % $\T \cup \censiga \modelseql \P$, where $\censiga=\bigcap_{\C\in\optcens(\E)}\C$.
    % \fi
    the set $\censiga=\bigcap_{\C\in\optcens(\E)}\C$ is a GA censor of $\E$.
\end{theorem}
\begin{forceproof}
    To be a GA censor of $\E$, $\censiga$ must be a subset of $\clt{\A}$ and be such that $\T\cup\censiga\modelseql\P$. The first condition always follows by construction of $\censiga$. Then, for both the cases have just to prove the second one.

    We first focus on full policies. To prove the thesis, let us consider any $\tau\in\P$ and any ground substitution $\sigma$ of the universal variables of $\tau$ such that $\T\cup\censiga\models\sigma(\body(\tau))$, and let us show that $\T\cup\censiga\models\sigma(\head(\tau))$. If $\T\cup\censiga\models\sigma(\body(\tau))$, then by monotonicity, it follows that $\T\cup\C\models\sigma(\body(\tau))$ for every $\C\in\optcens(\E)$.
    However, for every such $\C$, since $\tau$ is full, then $\sigma(\head(\tau))$ is a ground conjunction of atoms, and since $\C$ is optimal, then all such ground atoms belong to it. Consequently, all the ground atoms of $\sigma(\head(\tau))$ belong to $\censiga$, which immediately implies that $\T\cup\censiga\models\sigma(\head(\tau))$.

    We now turn our attention to linear policies. Notice that, if every ED is linear, then only one optimal GA censor $\C$ for $\E$ exists.
    Such a GA censor can indeed be deterministically computed by iteratively (until a fixpoint is reached) checking whether there exists an atom $\alpha\in\C$, an ED $\tau\in\P$ and an assignment $\sigma$ of the free variables of $\body(\tau)$ such that $\T\cup\{\alpha\} \models \sigma(\body(\tau))$ but $\T\cup\C \not\models \sigma(\head(\tau))$ and, in such a case, remove $\alpha$ from $\C$. 
    Thus, the intersection coincides with such a censor, i.e.\ $\censiga=\C$, which implies $\T\cup\censiga\modelseql\P$ by Definition~\ref{def:ga-censor}.
\end{forceproof}
    \section{Negative Results}
\label{sec:negative-results}

% \textcolor{red}{
% \begin{itemize}
%     \item \sout{è noto che GA-entailment è coNP-hard già nel caso dei denial (il che motiva la scelta della semantica IGA)} (solo accennato nella sezione~\ref{sec:framework}, per non introdurre un altro problema di decisione)
%     \item IGA-entailment è coNP-hard nel caso acyclic (non servirebbe se usassimo la sezione \ref{sec:security-of-intersection} per scartare il caso acyclic puro)
%     \item IGA-entailment è coNP-hard nel caso full
%     \item IGA-entailment è PTIME-hard nel caso linear
%     \item \sout{IGA-entailment è NL-hard nel caso di policy espresse in $\dlliter$} (dimostrato nella sez.~\ref{sec:experiments})
% \end{itemize}
% }

As anticipated in the Introduction, we aim to find a class of dependencies for which IGA-entailment is FO-rewritable.
Starting from the results obtained in the previous section, a natural first choice would be the classes of full and linear EDs. The following property, however, excludes the possibility of FO-rewritability of IGA-entailment for the class of linear epistemic dependencies.

% \begin{theorem}
% \label{thm:linear-iga-lb}
%     There exist an empty TBox, a policy consisting of linear EDs, and a query consisting of only one atom for which $\decProb$ is \PTIME-hard in data complexity.\nb{Lor: sostituire con il teorema~\ref{thm:binary-iga-lb}? È un bound più basso, ma è sufficiente per scartare la FO rewritability. Anche perché nell'articolo su rivista non abbiamo una vera e propria prova, ma sfruttiamo un teorema di Chomicki e Marcinkowski}
% \end{theorem}
% \begin{proof}
%     {\color{red} TODO}
% \end{proof}

\begin{theorem}
\label{thm:linear-iga-lb}
\label{thm:binary-iga-lb}
    There exist a policy consisting of linear EDs, and a query consisting of only one atom for which $\decProb$ is \NL-hard in data complexity, even in the case the TBox is empty.%\nb{Lor: ho riformulato così (idem per teorema seguente), perché un revisore ci ha fatto notare che ``There exist an empty TBox...'' suonava un po' strano.}
\end{theorem}
\begin{proof}
    \newcommand{\reachable}{\mathit{reachableVia}}
    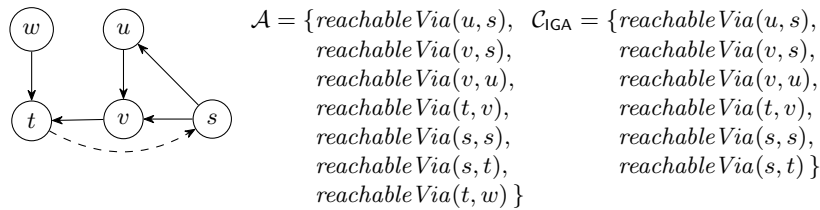
\begin{figure}[t]
        \centering
        \begin{minipage}[t]{0.25\textwidth}%
            \vspace{0pt}%
            \begin{tikzpicture}[vertex/.style={draw,circle}, ->, >={Stealth[round]}]
                \node[vertex] (w) [left] {$w$};
                \node[vertex] (u) [right=2em of w] {$u$};
                \node[vertex] (t) [below=2em of w] {$t$};
                \node[vertex] (v) [below=2em of u] {$v$};
                \node[vertex] (s) [right=2em of v] {$s$};
                
                \path
                    (w) edge (t)
                    (u) edge (v)
                    (v) edge (t)
                    (s) edge (v)
                    (s) edge (u)
                    (t) edge[bend right, dashed] (s);
            \end{tikzpicture}
        \end{minipage}
        \begin{minipage}[t]{0.7\textwidth}%
            \vspace{0pt}%
            \(
            \begin{array}{r@{}lr@{}l}
                \A =\{\,&\reachable(u,s),
                &\censiga =\{\,&\reachable(u,s),\\
                &\reachable(v,s),&&\reachable(v,s),\\
                &\reachable(v,u),&&\reachable(v,u),\\
                &\reachable(t,v),&&\reachable(t,v),\\
                &\reachable(s,s),&&\reachable(s,s),\\
                &\reachable(s,t),&&\reachable(s,t)\,\}\\
                &\reachable(t,w)\,\}&&
            \end{array}
            \)
        \end{minipage}
        \caption{A possible instance of the graph $\G$ and the corresponding sets $\A$ and $\censiga$, built as in the proof of Theorem~\ref{thm:binary-iga-lb}. The dashed edge only belongs to $\G'$.}
        \label{fig:reduction-graph}
    \end{figure}
    We prove the thesis by showing a logspace reduction from the restriction of st-connectivity to directed acyclic graphs (DAGs)\footnote{Such a problem is sometimes referred to as \textsc{DAG-STCON} or \textsc{STCONDAG}.}, which is known to be \NL-hard~\cite{M98}.
    Given a DAG $\G=\tup{V,E}$ and two nodes $s,t\in V$, such a problem consists of checking whether there exists a path from $s$ to $t$ in $\G$.
    
    First, let us fix the policy as follows:
    \[
    \P = \{ \forall u\,(K\exists v\,\reachable(v,u)\ra K\exists w\,\reachable(u,w)) \} .
    \]
    Now, let $\G'=\tup{V,E'}$ be such that $E'=E\cup\{(t,s)\}$, i.e.\ the graph identical to $\G$ except that it contains a further edge from $t$ to $s$.
    Let also $\T=\emptyset$ and
    \[
        \A = \{\reachable(v,u) \mid (u,v)\in E'\} \cup \{\reachable(s,s)\} .
    \]
    Note that $\A$ (we ignore the TBox, as it is empty) does not necessarily satisfy the policy, but the set $\{\reachable(s,s)\}$ does.
    Intuitively, the policy $\P$ states that, for every pair of nodes $v,u\in V$, $v$ is reachable from $s$ via $u$ only if $u$ is reachable from $s$ via another node $w\in V$.
    Also, the ABox states that every node is reachable from $s$ via all its predecessors in $\G'$ (which is not necessarily true), plus the fact that $s$ is reachable via itself.

    Notice also that, since the unique ED in $\P$ contains only one atom in its body, then there exists exactly one optimal GA censor for $\E=\tup{\T,\P,\A}$, which then coincides with $\censiga$.
    
    We can now prove that the following three statements are equivalent:
    \begin{enumerate}
        \item $\G$ contains a path from $s$ to $t$;
        \item there exists a directed cycle in $\G'$;
        \item the BCQ $q=\exists u\,\reachable(t,u)$ is IGA-entailed by $\E$.
    \end{enumerate}

    \medskip
    \noindent $(1 \Leftrightarrow 2)$ Since $\G$ is a DAG, it is immediate to verify that $\G$ contains a path from $s$ to $t$ iff $\G'$ is cyclic.

    \medskip
    \noindent $(2 \Rightarrow 3)$
    If a directed cycle exists in $\G'$, then it involves $s$ and $t$ (otherwise $\G$ would not be a DAG).
    In this case, $\A$ contains the following set of facts:
    \[\begin{array}{r@{\,}l}
        \{
        & \reachable(s,s), \reachable(u_1,s), \reachable(u_2,u_1),\\
        & \reachable(u_3,u_2), \ldots, \reachable(t,u_m), \reachable(s,t)
        \,\}
    \end{array}
    \]
    which %is disclosable in $\E$,\nb{Lor: forward reference!} and therefore it 
    is contained in the unique optimal GA censor for $\E$. Thus, we have that $\E\modelsiga q$.

    \medskip
    \noindent $(3 \Rightarrow 1)$
    Suppose that there is no path from $ s $ to $ t $ in $\G$, i.e.\ every directed path ending in $ t $ does not include $ s $. Also, since $\G$ is a DAG, none of these paths contains a cycle. Given this and the way $\A$ is constructed, it follows immediately that no fact of the form  $\reachable(t,\_)$ can be part of the unique optimal GA censor for $\E$ (which, as stated above, coincides with $\censiga$). Indeed, if any fact $\reachable(t,u)$ were included in $\censiga$, the policy would require the inclusion of another fact $\reachable(u,w)$, and so on, until one eventually reaches a fact (corresponding to the final edge of a path ending at $ t $) beyond which no further fact can be added. Therefore, we have that $\E\not\modelsiga q$.
    \qed
\end{proof}

\medskip
We now show that the scenario is even worse for full dependencies, as the decision problem under consideration is even intractable in data complexity.

\begin{theorem}
\label{thm:full-iga-lb}
    There exist a policy consisting of full EDs, and a query consisting of a ground atom for which $\decProb$ is \coNP-hard in data complexity, even in the case the TBox is empty.
\end{theorem}
\begin{proof}
    We prove the thesis by reduction from 3-CNF.
    We define the TBox $\T=\emptyset$ and the following full policy:
    \[
    \begin{array}{r@{\ }l}
        \forall i,j,p,v & (K\,(S(i)\wedge
        N(j,i)\wedge V_1(i,p)\wedge P_1(i,v)\wedge T(p,v)) \rightarrow K\,S(j)) \\
        \forall i,j,p,v & (K\,(S(i)\wedge
          N(j,i)\wedge V_2(i,p)\wedge P_2(i,v)\wedge T(p,v)) \rightarrow K\,S(j)) \\
        \forall i,j,p,v & (K\,(S(i)\wedge
         N(j,i)\wedge V_3(i,p)\wedge P_3(i,v)\wedge T(p,v)) \rightarrow K\,S(j)) \\
        \forall x & (K\,(T(x,0)\wedge T(x,1)) \rightarrow K\,U(0))
        % K\,S(0) \rightarrow K\,U
    \end{array}
    \]
    %
    % Given a 3-CNF formula $\varphi$ with $m$ clauses, in the ABox $\A$ we represent every clause of $\varphi$ using the binary predicates $N$ (indicating that a clause directly succeeds another one), $V_j$ and $P_j$ with $j\in\{1,2,3\}$ (indicating, respectively, the position of the $j$-th variable and the polarity of the $j$-th literal).
    % %%
    % For example, if the $i$-th clause clause is of the form $a\lor\neg b\lor c$, then we add to $\A$ the facts $N(i-1,i)$, $V_1(i,a)$, $V_2(i,b)$, $V_3(i,a)$, $P_1(i,1)$, $P_2(i,0)$, $P_3(i,1)$.
    % For example, if the $i$-th clause clause is of the form $a\lor\neg b\lor c$, then we add to $\A$ the facts $N(i-1,i)$, $V_1(i,a)$, $V_2(i,b)$, $V_3(i,a)$, $P_1(i,1)$, $P_2(i,0)$, $P_3(i,1)$.
    % Moreover, the ABox contains the facts $T(p,0)$ and $T(p,1)$ for every propositional variable $p$, and the facts $\{ S(1),S(2),\ldots,S(m) \}$.
    %
    \iflong
    Now, given a 3-CNF formula $\varphi$ with $m$ clauses, we build the ABox $\A$ as the set of the following:
    \else
    Now, given a 3-CNF formula $\varphi$ with $m$ clauses, the ABox $\A$ contains the facts:
    \fi
    \begin{itemize}
        \item $N(i-1,i)$ for every $1\le i \le m$, meaning that $c_i$ is the next clause of $c_j$ (note that $N(0,1)$ is included);
        \item $V_j(i,p)$ if $p$ is the $j$-th propositional variable in the $i$-th clause;
        \item $P_j(i,0)$ (resp., $P_j(i,1)$) if the polarity of the $j$-th literal in the $i$-th clause is negative (resp., positive);
        \item $T(p,0)$ and $T(p,1)$ for every propositional variable $p$ occurring in $\varphi$;
        \item $S(1),S(2),\ldots,S(m)$.
    \end{itemize}
    
     Intuitively, the last ED affirms that every propositional variable can assume only one boolean value, and the first three EDs affirm that if a clause is satisfied (thus at least one of its three literals is true) the previous clause has to be satisfied too. For instance, let $\phi = (a \lor \lnot b \lor c) \land (b \lor d \lor e)$, then $\A = $
    \begin{gather*}
        \{N(0,1), N(1,2),\\
        V_1(1, a), V_2(1, b), V_3(1, c),
        V_1(2, b), V_2(2, d), V_3(2, e),\\
        P_1(1, 1), P_2(1, 0), P_3(1, 1),
        P_1(2, 1), P_2(2, 1), P_3(2, 1),\\
        T(a, 0), T(a, 1), T(b, 0), T(b, 1), T(c, 0), T(c, 1), T(d, 0), T(d, 1), T(e, 0), T(e, 1),\\
        S(1), S(2)
        \}
    \end{gather*}

    We prove that $\varphi$ is unsatisfiable iff $S(m)$ is IGA-entailed by $\E$.
    
    $(\Leftarrow)$ Suppose $\varphi$ is satisfiable. Let $P$ be the set of propositional variables occurring in $\varphi$, let $I\subseteq P$ be an interpretation satisfying $\varphi$, and let $\A'$ be the following subset of $\A$:
    \[
    \begin{array}{r@{}l}
    \A' = \A \setminus (
        &\{ T(p,0) \mid p\in I \}\;\cup\\
        &\{ T(p,1) \mid p\in P\setminus I \} \cup \{S(1),\ldots,S(m) \} )
    \end{array}
    \]
    
    It is immediate to verify that $\T\cup\A'\modelseql\P$, as it does not contain any $S$ fact and there does not exist any constant $c$ such that $\{T(c,0),T(c,1)\}\subseteq\A'$.
    This implies that $\A'$ is part of at least one optimal GA censor of $\E$ (as it is a GA censor of $\E$ itself).
    Moreover, one can see that $\A'\cup \{S(m)\}$ is not part of any optimal GA censors of $\E$. This is proved by the fact that the addition of $S(m)$ creates a sequence of instantiations of the bodies of the first three EDs of $\P$ that requires (to preserve the satisfaction of the policy) to add to $\A'\cup\{S(m)\}$ first the fact $S(m-1)$, then $S(m-2)$, and so on until $S(1)$: this would in turn imply to also add $S(0)$, which however, does not belong to $\clt{\A}$.
    
    Consequently, there exists an optimal GA censor $\C$ of $\E$ that does not contain $S(m)$. Thus, $S(m)$ is not part of the intersection $\censiga$ of the optimal GA censors%. From this, we directly have that $\T\cup\censiga\not\models S(m)$, i.e.\ $\E\not\modelsiga S(m)$. Indeed, in case $\T\cup\censiga\models S(m)$, by monotonicity we would also have that $\T\cup\C\not\models S(m)$, and since $\C$ is optimal we would have that $S(m)\in\C$, which is a contradiction.
    , and since the TBox $\T$ is empty, we directly have that $\T\cup\censiga\not\models S(m)$, i.e.\ $\E\not\modelsiga S(m)$.
    
    $(\Rightarrow)$ Given a guess of the atoms of the $T$ predicate satisfying the fourth dependency and corresponding to an interpretation of the propositional variables that does not satisfy $\varphi$, it is straightforward to verify that the sequence of instantiations of the bodies of the first three EDs of $\P$ mentioned above (which has previously lead to the need of adding $S(0)$ to the set) %is blocked by 
    results to be interrupted due to the absence of some fact for $T$. This missing fact reflects a variable whose truth value conflicts with the polarity of the corresponding literal.
    Then, there exists a positive integer $k\leq m$ such that the atoms $S(k), S(k+1), \ldots, S(m)$ can be added to all the optimal GA censors corresponding to such a guess of the $T$ atoms. Since this holds for every interpretation of $\varphi$ (none of which satisfies the propositional formula), then the unsatisfiability of $\varphi$ implies that $\E\modelsiga S(m)$.
    \qed
\end{proof}

%%%%%%%%%%%%%%%%%%%%%%%%%%%%%%%%%%%%%%%%%

    \section{A FO-Rewritable Class of EDs}
\label{sec:fo-rewritability}

In this section, we identify a condition for full EDs such that it is possible to solve IGA-entailment through an FO-rewriting technique in the case of $\dlliter$ TBoxes.
We start by defining when a given set $\F$ of facts is disclosable. Intuitively, this happens when we can expose it via at least one optimal GA censor.

\begin{definition}[Disclosability]
\label{def:disclosability}
    Given a CQE instance $\E$, we say that a set of facts $\F$ is \emph{disclosable in $\E$} if there exists a subset $\F'$ of $\clt{\A}$ such that $\F\subseteq\F'$ and $\T\cup\F'\modelseql\P$.
    % Given a CQE instance $\E$, we say that a set of facts $\F$ is \emph{disclosable in $\E$} if there exists a GA censor $\C$ for $\E$ such that $\F\subseteq\C$.
\end{definition}

\begin{algorithm}[t]
\caption{$\policyexp$}
\label{alg:policy-expand}
    \Input{A $\dlliter$ TBox $\T$, a policy $\P$ expandable w.r.t.\ $\T$}
    %$\Sigma\gets\TGD(\P^+)\cup\TGD(\T)$;\\
    $\P'\gets\emptyset$;\\
    \ForEach{$\tau\in\P$}{
        \ForEach{$q(\vseq{x})\in\ucqrew(\body(\tau),\Sigma(\P,\T))$}{
            %\textbf{remove} all the $\ind$ atoms from $q(\vseq{x})$;\\
            $\P'\gets\P'\cup\{\forall\vseq{x} (K q(\vseq{x}) \ra K \head(\tau))\}$;
        }
    }
    \Return{$\P'$}
\end{algorithm}

For technical purposes, we make use of the notion of \emph{tuple-generating dependency} (TGD)~\cite{AbHV95}, i.e.\ FO expressions of the form $\forall\vseq{x}_1,\vseq{x}_2\,(q_b(\vseq{x}_1,\vseq{x}_2)\ra q_h(\vseq{x}_2))$, where both $q_b(\vseq{x}_1,\vseq{x}_2)$ and $q_h(\vseq{x}_2)$ are CQs.
We say that a set $\Sigma$ of TGDs is \emph{UCQ-rewritable} if, given any CQ $q(\vseq{x})$, there exists a UCQ $\ucqrew(q,\Sigma)$ such that, for every set $\F$ of facts and for every ground substitution $\sigma$ of the free variables of $q$, $\Sigma\cup\F\models \sigma(q)$ iff $\F\models\sigma(q_r)$ for some $q_r(\vseq{x})\in\ucqrew(q,\Sigma))$.
In this regard, we refer to~\cite{BLMS11}, which establishes a sufficient condition for UCQ rewritability (generalizing acyclicity and linearity), and to~\cite{KLMT15}, which provides a suitable rewriting algorithm% that we here indicate as $\ucqrew$
.\footnote{Actually, that paper focuses on BCQs. Anyway, as also stated by the authors, the algorithm provided is easily extendable to open CQs.}

Now, given a policy $\P$, let us denote by $\TGD(\P)$ the following set of TGDs:
\[
\{
    \forall\vseq{x}_1,\vseq{x}_2\,(q_b(\vseq{x}_1,\vseq{x}_2)\ra q_h(\vseq{x}_2)) \mid
    \forall\vseq{x}_1,\vseq{x}_2\,(K q_b(\vseq{x}_1,\vseq{x}_2)\ra K q_h(\vseq{x}_2)) \in\P\,
\} .
\]
Moreover, given a $\dlliter$ TBox $\T$, let us denote by $\TGD(\T)$ the set of TGDs obtained from $\T$ in the natural way. 
\ifcameraready\else More details are given in the Appendix.\fi

UCQ-rewritability is a well-established property of $\dlliter$ TBoxes~\cite{CDLLR07}\iflong, i.e.\fi:
\begin{proposition}
\label{pro:dlliter-ucqrew}
    Let $\T$ be a $\dlliter$ TBox. Then, $\TGD(\T)$ is UCQ-rewritable.
\end{proposition}
For the sake of readability, when referring to $\dlliter$ TBoxes $\T$, we write $\ucqrew(q,\T)$ instead of $\ucqrew(q,\TGD(\T))$.

Furthermore, let us indicate with $\P^+$ the policy containing every ED in $\P$ whose head does not contain $\bot$, and with $\Sigma(\P,\T)$ the set of TGDs $\TGD(\P^+)\cup\TGD(\T)$. We say that $\P$ is \emph{expandable} w.r.t.\ $\T$ if $\Sigma(\P,\T)$ is UCQ-rewritable. Then, from the results in~\cite{KLMT15}, we directly have what follows:
\begin{proposition}
\label{pro:lin-acy-ucqrew}
    Let $\T$ be a $\dlliter$ TBox, and let $\P$ be a policy that is either linear or acyclic for $\T$. Then, $\P$ is expandable w.r.t.\ $\T$.\footnote{It is not hard to see that this property is preserved in the case the policy also contains denials. Thus, our technique applies to a generalization of the case of denials.}
\end{proposition}
% \begin{forceproof}
%     It is immediate to verify that, if $\P$ is linear (resp., acyclic), then the set $\Sigma(\P,\T)$ is a linear (resp., acyclic in the classical sense) set of TGDs.
% \end{forceproof}

\iflong 
We can now introduce Algorithm~\ref{alg:policy-expand}, which takes as input a $\dlliter$ TBox $\T$ and a policy $\P$ expandable w.r.t.\ $\T$ and returns a new policy $\P'$ obtained in the following way.
For each ED $\tau\in\P$, it builds one new ED for each way of rewriting the body of $\tau$ w.r.t.\ $\Sigma(\P,\T)$; in particular, it uses every BCQ $q(\vseq{x})$ occurring in the rewriting as the body of the new ED and it leaves unchanged the head, while universally quantifying the free variables of $q$. Note that every such ED has the same universally quantified variables of the original ED $\tau$.

% \begin{example}
%     {\color{red} TODO}
% \end{example}

Algorithm~\ref{alg:policy-expand} enjoys the following important property.
\else
We can now introduce Algorithm~\ref{alg:policy-expand}, for which the following property holds.
\fi
\begin{lemma}
\label{lem:policyexp}
    %Let $\T$ be a $\dlliter$ TBox, let $\P$ be a full and UCQ-rewritable policy, and let $\F$ be a set of facts. Then, $\F\modelseql\policyexp(\P,\T)$ iff $\T\cup\F\modelseql\P$.
    Let $\E=\tup{\T,\P,\A}$ be a $\dlliter$ CQE instance \iflong such that \else s.t.\ \fi $\P$ is full and expandable w.r.t.\ $\T$, and let $\F$ be a set of facts. Then, $\F$ is disclosable in $\E$ iff\iflong\ both the following conditions hold\fi:
    \begin{itemize}
        \item[$(i)$] $\F\subseteq\clt{\A}$\iflong;\else\ and\fi
        \item[$(ii)$] for every $\tau\in\policyexp(\P,\T)$ and every ground substitution $\sigma$ of the free variables of $\body(\tau)$, $\F\models\sigma(\body(\tau))$ implies that $\T\cup\A\models\sigma(\head(\tau))$.
    \end{itemize}
\end{lemma}
\begin{proof}
    $(\Rightarrow)$
    Suppose that $\F$ is disclosable in $\E$. By Definition~\ref{def:disclosability}, this means that there exists a subset $\F'$ of $\clt{\A}$ such that $\F\subseteq\F'$ and $\T\cup\F'\modelseql\P$.
    Then, condition $(i)$ directly follows from the fact that $\F\subseteq\F'\subseteq\clt{\A}$.
    
    Let us now consider any $\tau$ and $\sigma$ as in condition $(ii)$, and let $\F\models\sigma(\body(\tau))$. To prove the \emph{only-if} direction of the thesis, it remains to show that $\T\cup\A\models\sigma(\head(\tau))$.
    First, by construction of $\policyexp(\P,\T)$, there exists an ED $\tau'\in\P$ such that $\body(\tau)\in\ucqrew(\body(\tau'),\Sigma))$ (where $\Sigma$ is as in Algorithm~\ref{alg:policy-expand}) and $\head(\tau')=\head(\tau)$. Consequently, we have that $\F\models\ucqrew(\body(\tau'),\Sigma))$ and, since $\Sigma$ is UCQ-rewritable, that $\Sigma\cup\F\models\sigma(\body(\tau'))$.
    
    Now, by construction of $\F'$ and since $\P^+$ is full, for every BCQ $q$, we have that $\T\cup\TGD(\P^+)\cup\F\models q$ only if $\T\cup\F'\models q$.
    In particular, $\Sigma\cup\F\models\sigma(\body(\tau'))$ implies that $\T\cup\F'\models\sigma(\body(\tau'))$.
    Consequently, since $\T\cup\F'\modelseql\P$, we have that $\T\cup\F'\models\sigma(\head(\tau'))$. But since $\F'\subseteq\clt{\A}$, then by monotonicity it follows that $\T\cup\clt{\A}\models\sigma(\head(\tau'))$, which holds iff $\T\cup\A\models\sigma(\head(\tau'))$.
    Finally, from the fact that $\head(\tau')=\head(\tau)$, we have that $\T\cup\A\models\sigma(\head(\tau))$.
    
    \smallskip
    
    $(\Leftarrow)$
    Let now $\F$ be a set of facts satisfying conditions $(i)$ and $(ii)$.
    
    Let us take the sequences of EDs $\tau_1,\ldots,\tau_m$, substitutions $\sigma_1,\ldots,\sigma_m$ and sets of facts $\F_1,\ldots,\F_{m+1}$ (with $\F_1=\F$) such that, for every $1\le i\le m$:
    \begin{itemize}
        \item $\tau_i\in\P$;
        \item $\sigma_i$ is a ground substitution of the free variables of $\body(\tau_i)$ such that $\F_i\models\sigma_i(\body(\tau_i))$;
        \item $\F_{i+1}=\F_i\cup\atoms(\sigma_i(\head(\tau_i)))$.
    \end{itemize}
    %%
    %Since $\P$ is full, then 
    %Note that $m$ can be upper-bounded in such a way that $\F_{m+1}=\F_{m}$ independently on how $\tau_{i}$ and $\sigma_{i}$ are chosen. In particular, this happens when $\F_m$ consists of all the facts that are logical consequences of $\TGD(\P)\cup\F$.
    Note that, since $\P$ is full, it is possible to choose $m$ as a finite positive integer such that $\F_m$ consists of all the facts that are logical consequences of $\TGD(\P)\cup\F$.
    Let us then set $m$ to this upper bound.

    We now show by induction that $\F_i\subseteq\clt{\A}$ for every $1\le i\le m$.
    To this aim, let $\P_0=\emptyset$ and, for every $1\le i\le m$, let $\P_i$ be the set $\P_{i-1}\cup\{\tau_i\}$.
    Observe that, by construction, we have that $\TGD(\P_{i-1})\cup\F\models\sigma_i(\body(\tau_i))$ for every $1\le i\le m$.
    \begin{itemize}
        \item[$\lozenge$]
        The base case is trivial, as $\F_1=\F$ is contained in $\clt{\A}$ by condition $(i)$.
        
        \item[$\lozenge$]
        For the inductive step, suppose that $\F_{i-1}\subseteq\clt{\A}$.
        We first show that $\tau_j\in\P^+$ for every $1\le j\le i$. Towards a contradiction, let us consider the lowest $j$ for which $\tau_j$ is a denial (i.e.\ $\P_{j-1}=\P_{j-1}^+$).
        In this case, we would have that $\TGD(\P_{j-1}^+)\cup\F\models\sigma_j(\body(\tau_j))$, implying by monotonicity that $\TGD(\P^+)\cup\F\models\sigma_j(\body(\tau_j))$, and hence that $\Sigma\cup\F\models\sigma_j(\body(\tau_j))$ (where $\Sigma$ is as in Algorithm~\ref{alg:policy-expand}).
        Since $\Sigma$ is UCQ-rewritable, then there would exist a CQ $q\in\ucqrew(\body(\tau_j),\Sigma)$ having the same free variables of $\body(\tau_j)$ such that $\F\models\sigma_j(q)$.
        Moreover, by Algorithm~\ref{alg:policy-expand}, $\policyexp(\P,\T)$ contains the ED $\tau=\forall\vseq{x}\,(K q(\vseq{x})\ra K \head(\tau_j))$.
        Then, condition $(ii)$ would imply that $\T\cup\A\models\sigma_j(\head(\tau_j))$, i.e.\ $\T\cup\A\models\bot$, which would contradict the hypothesis that the ontology $\T\cup\A$ is consistent.
        
        From the fact that $\tau_j\in\P^+$ for every $1\le j\le i$ it follows that $\TGD(\P^+)\cup\F\models\sigma_i(\body(\tau_i))$. Thus, following the same reasoning as above, one can conclude $\T\cup\A\models\sigma_i(\head_i)$ (i.e.\ $\sigma_i(\head_i)\subseteq\clt{\A}$), which, together with the inductive hypothesis, implies that $\F_i\subseteq\clt{\A}$.
    \end{itemize}
    
    Finally, in order to prove that $\T\cup\F_m\modelseql\P$, let us take any ED $\tau\in\P$ and any ground substitution $\sigma$ of the free variables of $\tau$, and let us show that if $\T\cup\F_m\models\sigma(\body(\tau))$ then $\T\cup\F_m\models\sigma(\head(\tau))$.
    Suppose then that $\T\cup\F_m\models\sigma(\body(\tau))$, which by monotonicity holds only if $\Sigma\cup\F_m\models\sigma(\body(\tau))$.
    Since $\Sigma$ is UCQ-rewritable, there exists a CQ $q\in\ucqrew(\body(\tau),\Sigma)$, with the same free variables as $\body(\tau)$, such that $\F_m\models\sigma(q)$.
    Notice that $\policyexp(\P,\T)$ contains the ED $\tau'=\forall\vseq{x}\,(K q(\vseq{x})\ra K \head(\tau))$.
    Then, by the construction of $\F_m$, from $\F_m\models \sigma(q)$ (that is, $\F_m\models\sigma(\body(\tau'))$) it follows that $\atoms(\sigma(\head(\tau')))\subseteq\F_m$% (i.e.\ $\F_m\models\sigma(\head(\tau'))$)
    , which in turn implies that $\T\cup\F_m\models\sigma(\head(\tau'))$ and, since $\head(\tau')=\head(\tau)$, we conclude that $\T\cup\F_m\models\sigma(\head(\tau))$.
    
    Thus, $\F_m$ is the set that meets the two conditions of Definition~\ref{def:disclosability}, that is, $\F$ is disclosable in $\E$.
    \qed
\end{proof}

In the following, given a CQ $q$ and a set of atoms $\Z$, we say that $q$ is \emph{mappable to $\Z$ via $\mu$} if there exists a substitution $\mu$ replacing the variables of $\Z$ and the free variables of $q$ with terms of $\Z$ and constants of $q$,
% \begin{itemize}
%     \item the variables of $\Z$ with terms of $\Z$ and constants of $q$ and
%     \item the free variables of $q$ with terms of $\Z$
% \end{itemize}
in such a way that there exists a substitution $\mu'$ of the existential variables of $q$ such that $\mu'(\mu(\atoms(q)))\subseteq\mu(\Z)$.
We indicate as $\map(q,\Z)$ the set of all the substitutions $\mu$ such that $q$ is mappable to $\Z$ via $\mu$. %and are most general, i.e.\ there does not exist any $\mu'$ such that $q$ is mappable to $\Z$ and $\mu$ specializes $\mu'$.\nb{Lor: va bene come definizione di "most general"?}
E.g.\ if $\Z=\{R(x,y),R(z,w)\}$ and $q(v,u)=\exists t\,R(v,1)\land R(u,t)$, then both $\{v\mapsto x,y\mapsto 1,u\mapsto z\}$ and $\{v\mapsto z,w\mapsto 1,u\mapsto x\}$ belong to $\map(q,\Z)$.

\begin{definition}\label{def:isdiscl-formula}
    Given a $\dlliter$ TBox $\T$, a policy $\P$ expandable w.r.t.\ $\T$, and a set $\Z$ of atoms  (we assume w.l.o.g.\ that every $x\in\vars(\Z)$ does not occur in $\policyexp(\P,\T)$), we define the formula $\isdisclformula(\Z,\T,\P)$ as follows:
    \[
        \atomrewr(\conj(\Z),\T) \land 
        \bigwedge_{\substack{
            \tau \in \policyexp(\P,\T)\;\land\\
            \mu \in \map(\body(\tau),\Z)
        }}
        \Big(
        \eq(\mu,\Z) \ra \ucqrew(\mu(\head(\tau)),\T)
        \Big)
    \]
    where:
    \begin{itemize}
        \item $\conj(\Z) = \bigwedge_{\alpha\in\Z}\alpha$, for every set $\Z$ of atoms;
        \item $\displaystyle\eq(\mu,\Z) = \bigwedge_{\substack{x \mapsto t \in \mu\;\land\\ x \in\vars(\Z)}} x=t$ (with $true$ in place of the empty conjunction).
        %\nb{Verificare se l'implementazione è in accordo alla definizione, perché Flavia quantifica su tutti i "padding", cioè tutti i sottoinsiemi di una versione senza unificazione dei body (smart), e poi considera dentro IsDiscl sono i mapping che usano l'intero padding per formare un clash insieme alla query. Update: non funziona, perché bisogna comunque verificare i sottoinsiemi del padding potenzialmente non-disclosable (si potrebbe limitare il check solo ai sottoinsiemi che hanno predicati sufficienti ad attivare da soli un body di ED).}
    \end{itemize}
\end{definition}

Intuitively, for every ground substitution $\sigma$ of its free variables, the FO sentence $\sigma(\isdisclformula(\Z,\T,\P))$ encodes the two conditions of Lemma~\ref{lem:policyexp}. \iflong This property is formally established by the next statement. \else More formally: \fi

\begin{lemma}
\label{lem:discl-rewriting}
    Let $\E=\tup{\T,\P,\A}$ be a $\dlliter$ CQE instance such that $\P$ is full and expandable w.r.t.\ $\T$.
    Then, for every set of atoms $\Z$ and for every ground substitution $\sigma$ of the variables of $\Z$,
    $\sigma(\isdisclformula(\Z,\T,\P))$ evaluates to true in $\A$ iff $\sigma(\Z)$ is disclosable in $\E$.
\end{lemma}
\begin{proof}
    $(\Leftarrow)$
    Let $\sigma(\Z)$ be disclosable in $\E$. By Lemma~\ref{lem:policyexp}, we have that:
    \begin{enumerate}
        \item[$(i)$] $\sigma(\Z)\subseteq\clt{\A}$ and
        \item[$(ii)$] for every $\tau\in\policyexp(\P,\T)$ and every ground substitution $\sigma'$ of the free variables of $\body(\tau)$, $\sigma(\Z)\models\sigma'(\body(\tau))$ implies that $\T\cup\A\models\sigma'(\head(\tau))$.
    \end{enumerate}
    Clearly, from condition $(i)$ it follows that $\sigma(\conj(\Z))$ evaluates to true in $\clt{\A}$. By Proposition~\ref{pro:atomrewr}, we then have that $\atomrewr(\sigma(\conj(\Z)),\T)$ (which coincides with $\sigma(\atomrewr(\conj(\Z),\T))$) evaluates to true in $\A$.
    
    Now, let us consider any ED $\tau\in\policyexp(\P,\T)$ and any substitution $\mu\in\map(\body(\tau),\Z)$.
    By definition, $\mu$ can be partitioned in two disjoint substitutions, namely $\mu_1$ and $\mu_2$, such that:
    \begin{itemize}
        \item $\mu_1$ is a substitution of the free variables of $\body(\tau)$ with terms of $\Z$ and constants of $\body(\tau)$ such that $\mu(\body(\tau))=\mu_1(\body(\tau))$;
        \item $\mu_2$ is a substitution of the variables of $\Z$ with terms of $\Z$ and constants of $\body(\tau)$ such that $\mu(\Z)=\mu_2(\Z)$;
        \item there exists a substitution $\mu'$ of the existential variables of $\body(\tau)$ (i.e.\ the ones of $\mu_1(\body(\tau))$) such that $\mu'(\mu_1(\atoms(\body(\tau))))\subseteq\mu_2(\Z)$.
    \end{itemize}
    
    Consider now the conjunction of atoms $\sigma(\eq(\mu,\Z))$, which, by definition of $\eq(\cdot,\cdot)$, is equal to $\sigma(\eq(\mu_2,\Z))$. As it is ground, it can either be unsatisfiable or valid (i.e.\ a conjunction of reflexive equalities on constants). In the first case, it obviously evaluates to false in $\A$.
    In the second case, instead, we have that %every equality $\lambda$ induced by $\mu_2$ is such that $\sigma(\lambda)$ is %of the form $c=c$ for some constant $c$
    %a tautology, which is possible only if $\sigma(\mu_2(\Z))=\sigma(\Z)$.
    by applying $\sigma$ to any variable $x$ of $\Z$ or to the term which $x$ is mapped to via $\mu_2$ one obtains the same constant, which implies that $\sigma(\mu_2(\Z))=\sigma(\Z)$.
    Moreover, since $\sigma$ does not replace existential variables of $\mu_1(\atoms(\body(\tau)))$, then by construction of $\mu'$ we have that $\mu'(\sigma(\mu_1(\atoms(\body(\tau)))))\subseteq\sigma(\mu_2(\Z))$, which holds only if $\sigma(\Z)\models\sigma(\mu_1(\body(\tau)))$.
    Now, note that the substitution $\sigma'$ resulting by applying first $\mu_1$ and then $\sigma$ to $\body(\tau)$ is a ground substitution of the free variables of $\body(\tau)$.
    Therefore, by condition $(ii)$, it follows that $\T\cup\A\models\sigma(\mu_1(\head(\tau)))$ (or equivalently $\T\cup\clt{\A}\models\sigma(\mu_1(\head(\tau)))$). Then, by Proposition~\ref{pro:dlliter-ucqrew}, the sentence $\sigma(\ucqrew(\mu_1(\head(\tau)),\T))$ (i.e.\ $\sigma(\ucqrew(\mu(\head(\tau)),\T))$) evaluates to true in $\A$.
    
    Thus, for every $\tau\in\policyexp(\P,\T)$ and for every $\mu\in\map(\body(\tau),\Z)$, either $\sigma(\eq(\mu,\Z))$ evaluates to false in $\A$ or $\sigma(\ucqrew(\mu(\head(\tau)),\T))$ evaluates to true in $\A$, from which the thesis immediately follows.
    
    \smallskip
    
    $(\Rightarrow)$
    Now, suppose that $\sigma(\isdisclformula(\Z,\T,\P))$ evaluates to true in $\A$. Then:
    \begin{enumerate}
        \item[$(i)$] $\sigma(\atomrewr(\conj(\Z),\T))$ evaluates to true in $\A$, which by Proposition~\ref{pro:atomrewr} implies that $\sigma(\conj(\Z))$ evaluates to true in $\clt{\A}$ (i.e.\ $\sigma(\Z)\subseteq\clt{\A}$);
        \item[$(ii)$] for every $\tau\in\policyexp(\P,\T)$ and for every $\mu\in\map(\body(\tau),\Z))$, if the (ground) conjunction $\sigma(\eq(\mu,\Z))$ evaluates to true in $\A$ then also the BUCQ $\sigma(\ucqrew(\mu(\head(\tau)),\T))$ does.
    \end{enumerate}
    
    Let us consider any ED $\tau\in\policyexp(\P,\T)$ and any ground substitution $\sigma'$ of the free variables of $\body(\tau)$ such that $\sigma(\Z)\models\sigma'(\body(\tau))$. For proving the thesis, we have to show that $\T\cup\A\models\sigma'(\head(\tau))$, which by Lemma~\ref{lem:policyexp} would imply that $\sigma(\Z)$ is disclosable in $\E$.

    First, since $\sigma'$ replaces the free variables of $\body(\tau)$ with constants of $\sigma(\Z)$, then there exists a substitution $\mu_1$ of the free variables of $\body(\tau)$ with terms of $\sigma(\Z)$ such that $\sigma'(\body(\tau))=\sigma(\mu_1(\body(\tau))$. Therefore, it holds that $\sigma(\Z)\models\sigma(\mu_1(\body(\tau)))$, i.e.\ $\sigma(\Z)\supseteq\mu'(\sigma(\mu_1(\atoms(\body(\tau)))))$ for some substitution $\mu'$ of the existential variables of $\sigma(\mu_1(\body(\tau)))$ (i.e.\ of the ones of $\body(\tau)$).
    Then, one can see that there exists a substitution $\mu_2$ of the variables of $\Z$ with terms of $\Z$ and constants of $\body(\tau)$ such that $\sigma(\Z)=\sigma(\mu_2(\Z))$ and $\mu_2(\Z)\supseteq\mu'(\mu_1(\atoms(\body(\tau))))$.
    Intuitively, $\mu_2$ is a weakened version of $\sigma$ that allows not to apply $\sigma$ on the right-hand side while preserving the homomorphic relationship between the two sets.
    
    Now, let $\mu_1$ and $\mu_2$ be two substitutions such that what above holds, and observe that they replace distinct variables. Then, it is easy to see that the combination $\mu$ of $\mu_1$ and $\mu_2$ belongs to $\map(\body(\tau),\Z)$.
    Moreover, since $\sigma(\Z)=\sigma(\mu_2(\Z))$, then we have that $\sigma(\eq(\mu_2,\Z))$ (which by construction is equivalent to $\sigma(\eq(\mu,\Z))$) is valid and, consequently, it evaluates to true in $\A$.
    Therefore, by condition $(ii)$, we have that $\sigma(\mu(\ucqrew(\head(\tau),\T)))$ evaluates to true in $\A$.
    By Proposition~\ref{pro:dlliter-ucqrew}, this implies that $\T\cup\A \models \sigma(\mu(\head(\tau)))$.
    Then, the thesis follows by observing that $\sigma(\mu(\head(\tau)))=\sigma(\mu_1(\head(\tau)))=\sigma'(\head(\tau))$.
    \qed
\end{proof}

Then, given a $\dlliter$ TBox $\T$, a policy $\P$ expandable w.r.t.\ $\T$, a set $\Z$ of atoms, and a CQ $q(\vseq{x})$ without existential variables (we assume w.l.o.g.\ that $\vseq{x} \cap \vars(\Z)=\emptyset$), we define the FO formula $\clashformula(\Z,q,\T,\P)$ as follows:
\begin{center}
    \(
    \clashformula(\Z,q,\T,\P) =
    \exists \vseq{y} \: \big(\isdisclformula(\Z,\T,\P) \wedge \neg \isdisclformula(\Z\cup\atoms(q),\T,\P) \big) ,
    \)
\end{center}
where $\vseq{y}$ is a tuple containing all the variables occurring in $\Z$. % and $\atoms(q)$ is the set of atoms occurring in $q$.
Note that the variables of $\vseq{x}$ are free.
In words, by existentially closing the formula $\clashformula(\Z,q,\T,\P)$ and then evaluating it over an ABox $\A$ it is possible to check whether there exists a common instantiation $\sigma$ for all the atoms of $\Z$ and $q$ such that $(i)$ the set $\sigma(\Z)$ is disclosable (and thus it is part of some censor) and $(ii)$ it is no longer disclosable when we add the atoms of $q$. Intuitively, if there exists an instantiation of the variables of $q$ for which the above does not occur for any $\Z$, then $q$ is entailed by $\E$ under the IGA semantics.

In order to obtain a proper FO-rewriting algorithm, it remains to show that the size of $\Z$ can be upper-bounded by a certain integer $k$ that is independent of $\A$.
To this aim, given a set of predicates $\pr=\{p_1,\ldots,p_m\}$ and a positive integer $k$, we define the following set of atoms:
\[
    \allatoms(\pr,k) =
    %\bigcup_{p\in\pr}
    %\bigcup_{i\in\{1,\ldots,k\}}
    %\{p(\vseq{x}_i)\},
    \{ p(\vseq{x}_i) \mid p\in\pr \mbox{ and } i\in\{1,\ldots,k\} \} ,
\]
where each $\vseq{x}_i$ is a sequence of $h$ fresh variables, if $h$ is the arity of $p$.

We are now ready to provide the following FO-rewriting function.
\begin{definition}
\label{def:iga-ent-sentence-full}
    Let $\T$ be a $\dlliter$ TBox, let $\P$ be a policy expandable w.r.t.\ $\T$, and let $q$ be a BUCQ. %, and let $k$ be a positive integer. 
    Then $\igaformula(q,\T,\P)$ is the sentence:
    \[
        \igaformula(q,\T,\P) = \bigvee_{\exists\vseq{x}\,\cnj(\vseq{x})\in q_r}
        \exists \vseq{x} \: \Big( \cnj\, \wedge \bigwedge_{\substack{\Z\subseteq\allatoms(\pred(\P\cup\T),k)\\\land |\Z|< k
        }} \neg \clashformula(\Z,\cnj,\T,\P) \Big) .
    \]
    where $q_r=\ucqrew(q,\T)$ and $k=\displaystyle\max_{\tau\in\policyexp(\P,\T)}|\atoms(\body(\tau))|$.
\end{definition}

%Note that $k=0$ in the case $\P$ is linear, which considerably simplifies the structure of the rewritten formula.\nb{Lor: non è vero se assumiamo che le policy lineari possano avere anche i denial!}

It is possible to prove that, for every $\dlliter$ CQE instance $\E=\tup{\T,\P,\A}$, the sentence $\igaformula(q,\T,\P)$ evaluates to true in $\A$ iff $\E\modelsiga q$. Such an FO-rewritability property implies the next theorem.

\begin{theorem}
\label{thm:iga-aczero}
    %Let $\E=\tup{\T,\P,\A}$ be $\dlliter$ CQE instance such that $\P$ is full and expandable w.r.t.\ $\T$, and let $q$ be a BUCQ. Deciding whether $\E$ IGA-entails $q$ is in $\aczero$ in data complexity.
    $\decProb$ is FO-rewritable and hence in $\aczero$ in data complexity in the case of $\dlliter$ TBoxes and policies that are full and expandable w.r.t.\ the coupled TBox.
\end{theorem}
\iflong\else
\begin{proofsk}
    Let $\phi$ be the $i$-th disjunct of $\igaformula(q,\P,\T)$ and let $\cnj$ be the BCQ occurring in its left-hand side.
    First, by Lemma~\ref{lem:discl-rewriting}, we can prove that $\sigma(\clashformula(\Z,cnj,\T,\P))$ evaluates to true in $\A$ iff there exists an instantiation $\sigma'$ of $\conj(\Z)$ in $\A$ such that $\sigma'(\Z)$ is disclosable in $\E$ and $\sigma'(\Z)\cup\sigma(\atoms(\cnj))$ is not.
    Then, we are able to show that the integer $k$ defined as in Definition~\ref{def:iga-ent-sentence-full} is sufficient for checking, in a sound and complete way, whether there exist a set $\Z$ and a substitution $\sigma$ s.t.\ $\sigma(\Z)$ is disclosable in $\E$ and $\sigma(\Z)\cup M$ is not, for every image $M$ of $\cnj$ in $\A$.
    This can easily lead us to prove that $\T\cup\censiga\models q$ iff $\igaformula(q,\T,\P)$ evaluates to true in $\A$, which in turn proves the thesis.
\end{proofsk}
\fi
\begin{proof}
    Let $\A$ be any ABox, let $q_i=\exists\vseq{x}\,\cnj(\vseq{x})$ be the BCQ such that $\cnj$ occurs in the $i$-th disjunct of $\igaformula(q,\T,\P)$, let $\Z$ be a set of atoms, and let $\sigma$ be an instantiation of $q_i$ in $\A$. Then, it is immediate to verify that  $\sigma(\clashformula(\Z,\cnj,\T,\P))$ evaluates to true in $\A$ iff there exists an instantiation $\sigma'$ of $\conj(\Z)$ in $\A$ such that $\isdisclformula(\sigma'(\Z),\T,\P)$ and $\isdisclformula(\sigma'(\Z)\cup\sigma(\atoms(q_i)),\T,\P)$ evaluate, respectively, to true and false in $\A$.
    Consequently, by Lemma~\ref{lem:discl-rewriting}, we have the following property:
    \begin{quote}
        \textbf{(PR1)}:
        %Let $\Z$ be a set of atoms, and let $\sigma$ be an instantiation of $q_i$ in $\A$. Then, 
        The sentence $\sigma(\clashformula(\Z,\gamma,\T,\P))$ evaluates to true in $\A$ iff there exists an instantiation $\sigma'$ of $\conj(\Z)$ in $\A$ such that $\sigma'(\Z)$ is disclosable in $\E$ and $\sigma'(\Z)\cup\sigma(\atoms(q_i))$ is not.
    \end{quote}
    The next statement follows from the previous one and from the fact that, for every set of facts $\F\subseteq\A$ of size not greater than a given integer $k'$, there exists a subset $\Z$ of $\allatoms(\pred(\P\cup\T),k')$ such that $\F$ is an image of $\conj(\Z)$ in $\A$.
    \begin{quote}
        \textbf{(PR2)}:
        Let $\phi$ be the $i$-th disjunct of $\igaformula(q,\P,\T)$. Then $\phi$ evaluates to true in $\A$ iff there exists 
        % an instantiation $\sigma$ of $\vars(q_i)$ 
        an image $M$ of $q_i$ in $\A$ such that, 
        for every set of facts $\F$ such that $\F\subseteq\A$ and $|\F|<k$ (where $k$ is as in Definition~\ref{def:iga-ent-sentence-full}), 
        if $\F$ is disclosable in $\E$ then also $\F\cup M$ is.
    \end{quote}
    Furthermore, the fact that $\P$ is expandable w.r.t.\ $\T$ implies that, if there exists a set $\F$ of facts such that $\F$ is disclosable in $\E$ and $\F\cup M$ is not, then there exists a set of facts $\F'$ such that $\F'$ is disclosable in $\E$, $\F'\cup M$ is not and $|\F'|<k$.
    
    This last property, along with (PR2), implies that $q_i$ evaluates to true in $\censiga$ iff the $i$-th disjunct of $\igaformula(q,\T,\P)$ evaluates to true in $\A$.
    Note in fact that, for every instantiation $\sigma$ of $q_i$ in $\A$, the above set $\F'$ of facts exists iff $\sigma(\atoms(q_i))$ is not contained in at least one optimal GA censor for $\E$ (i.e.\ the one containing $\F'$).
    Therefore, by Proposition~\ref{pro:dlliter-ucqrew} and since $q_i\in\ucqrew(q,\T)$, we have that $\T\cup\censiga\models q$ (i.e.\ $\E\modelsiga q$) iff $\igaformula(q,\T,\P)$ evaluates to true in $\A$, which proves that BUCQ entailment under IGA semantics is FO-rewritable, and thus in $\aczero$ \wrt data complexity.
    \qed
\end{proof}

\begin{example}
    Recalling Example~\ref{ex:iga-entailment}, let us verify that $\E\not\modelsiga q_1$ by rewriting it and then evaluating it over the ABox.
    First, observe that $\ucqrew(q_1,\T)=q_1$, i.e.\ the TBox does not affect the entailment of $q_1$ in this case, and that the unique instantiation of $q_1$ in $\A$ is $\sigma=\{x\mapsto\indA,y\mapsto\indB\}$.
    In addition, since $k=2$, all sets $\Z$ of Definition~\ref{def:iga-ent-sentence-full} are singletons. 
    In particular, for the set $\Z=\{\managerOf(x',y')\}$ one can see that, under the assignment $\sigma'=\sigma\cup\{x'\mapsto\indA,y'\mapsto\indB\}$% of the variables of $\Z\cup\atoms(q_1)$
    , the sentence $\sigma'(\clashformula(\Z,q_1,\T,\P))$ evaluates to true in $\A$. 
    Therefore, the whole sentence $\igaformula(q_1,\T,\P)$ evaluates to false in $\A$.
    
    For $q_2$, we have $\ucqrew(q_2,\T)=\{q_2,
    \exists x,z (\managerOf(x,z)\land\salary(x,\salaryA)),$ 
    $\exists x\,(\manager(x)\land\salary(x,\salaryA))\}$. 
    In particular, one can verify that for the last BCQ there exists no set $\Z'$ analogous to the above set $\Z$ for query $q_1$, which implies that $q_2$ is IGA-entailed by $\E$.
    \qedex
\end{example}
    \section{Experiments}
\label{sec:experiments}

\newcommand{\polAcy}{\P_\text{a}}
\newcommand{\polBin}{\P_\text{b}}
\newcommand{\polAcyRed}{\polAcy^\textbf{\hspace{1pt}-}}
\newcommand{\polBinRed}{\polBin^\textbf{\hspace{1pt}-}}

\newcommand{\benchFive}{\textsf{o2b}\textsubscript{5}\xspace}
\newcommand{\benchTen}{\textsf{o2b}\textsubscript{10}\xspace}

In this section, we describe the experiments that we conducted to test the feasibility of our approach.
We evaluated our queries on a standard laptop with an Intel i7\iflong-8565U\fi~@1.8~GHz processor and 16GB of RAM.

\newcommand\midtilde{\fontfamily{ptm}\selectfont\texttildelow}

We refer to the OWL2Bench benchmark for OWL ontologies~\cite{SBM20}, which models the university domain and includes a tool for generating ABoxes of customizable size (measured in number of universities). For our experiments, we tested all the ten SPARQL queries~\cite{owl2bench-queries} for \OWLQL against the \benchFive and \benchTen ABoxes, which store data about 5 and 10 universities (i.e.\ $\sim$325k and $\sim$710k ABox assertions), respectively. 
We removed the data properties from the input ontology (as they are not part of $\dlliter$), and added the definition of the \texttt{knows} object property and the \texttt{Woman} class in the TBox\footnote{\url{https://github.com/kracr/owl2bench/blob/master/UNIV-BENCH-OWL2QL.owl}} for \OWLQL, as both such predicates occur in the generated ABoxes.
The ABoxes were stored in an SQL database, as the FO query produced by our rewriting algorithm can be naturally translated into SQL.
We were thus able to delegate the evaluation of queries on the ABox to the SQL system.

As for the reasoner, we employed the tree-witness query rewriter for OWL2~QL ontologies\footnote{\url{https://titan.dcs.bbk.ac.uk/~roman/tw-rewriting/}}~\cite{KKZ12,MKZ13}. It can be used as an actual implementation of the $\ucqrew$ abstract rewriting function when its second argument is a $\dlliter$ TBox.

The two cases on which we focused our experiments are the ones in which the policy is either full and linear or full and acyclic for the coupled TBox. More precisely, for the first case we define a slightly more restricted language of EDs (defined below), which we call \emph{binary EDs}. The correspondence of such dependencies with $\dlliter$ axioms allows us to exploit the tree-witness rewriter in place of $\ucqrew$ for rewriting the EDs' bodies. For the case of acyclic policies, instead, we implemented a specific version of $\ucqrew$ from scratch.

% {\color{red} TODO: Descrivere ottimizzazioni per l'implementazione, ad es.:
% \begin{enumerate}
%     \item considerare solo le sostituzioni ``most general'' in $\map(q,\Z)$
%     \item selezionare i sottinsiemi di $\Z$ in base a qualche euristica
%     \item per ogni coppia di ED $K b \ra K h$ e $K b' \ra K\bot$, se $b\models b'$ (definire meglio) allora la prima ED si può rimuovere [NON FATTO]
%     \item Portare $\atomrewr(\Z)$ fuori da $\isdisclformula$
% \end{enumerate}
% }

Although Theorem~\ref{thm:iga-aczero} constitutes a remarkable theoretical result, from a practical point of view, the size of the rewritten formula $\igaformula$ may have a severe impact on the evaluation time. For this reason, we made several intensional optimizations in our implementation to obtain a simpler yet semantically equivalent rewriting. The most important are the following ones.
\begin{itemize}
    \item Instead of generating a $\clashformula$ subformula for every possible $\Z$ set defined in $\igaformula$, we can only consider those sets that, for a fixed $q\in q_r$, can match at least one body of the expanded policy.
    \item In the $\isdisclformula$ subformula, the atoms of the query can be omitted from $\conj(\Z)$ (at this point of the evaluation, we know that their rewriting is satisfied). Then, since the two $\atomrewr(\Z,\T)$ in $\clashformula$ become syntactically equal, they can only be evaluated once, that is, outside the two $\isdisclformula$.
    \item For its specific purpose, the output of the $\map$ function can be optimized in terms of specificity of the substitutions. Specifically, we do not use a substitution $\mu$ such that $q$ is mappable to $\Z$ via $\mu$ if we also use a substitution $\mu'$ that is more generic than $\mu$ and such that $q$ is mappable to $\Z$ via $\mu'$.
\end{itemize}

\iflong
Our system's source code can be downloaded from the URL: \url{https://github.com/anonymous-iswc25/iga-ed-rewriter/blob/main/iga-ed-rewriter.zip}.
The software required to run the evaluation tests is: Java 8+, MySQL, Python~3 (needed to create new ABoxes from scratch), and possibly Apache Maven (only to build the project from the source code).
\fi

\subsection{Binary EDs}

\newcommand{\DL}{\textit{DL}}

We now define the fragment of binary EDs, a subclass of linear EDs which has a correspondence with $\dlliter$.

\begin{definition}
\label{def:binary-eds}
    A \emph{binary ED} is an ED of one of the following forms:
    
    \iflong
    \begin{eqnarray}
    \forall x\, ( K B_1 \rightarrow K B_2 ) \label{eqn:binaryed-one} \\
    \forall x,y\, ( K S_1 \rightarrow K S_2 ) \label{eqn:binaryed-two} 
    \end{eqnarray}
    \else
    \smallskip
    \begin{minipage}{0.45\textwidth}
    \begin{equation}
      \forall x\, ( K B_1 \rightarrow K B_2 ) \label{eqn:binaryed-one}
    \end{equation}
    \end{minipage}
    \quad
    \begin{minipage}{0.45\textwidth}
    \begin{equation}
      \forall x,y\, ( K S_1 \rightarrow K S_2 ) \label{eqn:binaryed-two} 
    \end{equation}
    \end{minipage}
    
    \medskip
    \noindent
    \fi
    \iflong
    where $B_1$ and $B_2$ are expressions of one of the following forms:
    \[
    \{ A(x),\; \exists y\, R(x,y),\; \exists y\, R(y,x) \}
    \]
    (where $A$ is a concept name and $R_1,R_2$ are role names),
    and $S_1$ and $S_2$ are expressions of one of the following forms:
    \[
    \{ R(x,y),\; R(y,x) \}
    \]
    \else
    where $B_1,B_2\in\{ A(x), \exists y\, R(x,y), \exists y\, R(y,x) \}$, $S_1,S_2\in\{ R(x,y), R(y,x) \}$, $A\in\conceptSet$, and $R\in\roleSet$.
    \fi
\end{definition}
We now define the function $\DL(\cdot)$ as follows:
\[
\begin{array}{r@{\qquad}l@{\qquad}l}
    \DL(A(x)) = A
    & \DL(\exists y\, R(y,x)) = \exists R^-
    & \DL(R(x,y)) = R \\
    & \DL(\exists y\, R(x,y)) = \exists R
    & \DL(R(y,x)) = R^-
\end{array}
\]
% \[
% \begin{array}{r@{\qquad}l@{\qquad}l}
%     \DL(A(x)) = A
%     & \DL(R(x,y)) = R
%     & \DL(R(y,x)) = R^-
% \end{array}
% \]
Then, given a binary ED $\tau$ of the form (\ref{eqn:binaryed-one}), we define $\DL(\tau)$ as the $\dlliter$ concept inclusion $\DL(B_1)\ISA \DL(B_2)$, and given a binary ED $\tau$ of the form (\ref{eqn:binaryed-two}), we define $\DL(\tau)$ as the $\dlliter$ role inclusion $\DL(S_1)\ISA \DL(S_2)$.
Finally, if $\P$ is a set of binary EDs, we define $\DL(\P)$ as the $\dlliter$ TBox $\bigcup_{\tau\in\P} \DL(\tau)$.

% \bigskip

% Given a set of binary EDs $\P$ and a $\dlliter$ TBox $\T$, we define $\policyexp(\P,\T)$ as the following set of $\dlliter$ inclusions:
% \[
% \begin{array}{l}
% \policyexp(\P,\T) = \policyexp_c(\P,\T) \cup \policyexp_r(\P,\T) \\
% \textrm{where}\\
% \policyexp_c(\P,\T) = \{ B_1'\ISA B_2 \mid B_1\ISA B_2 \in \P \wedge \T\models B_1'\ISA B_1 \} \\
% \policyexp_r(\P,\T) = \{ S_1'\ISA S_2 \mid S_1\ISA S_2 \in \P \wedge \T\models S_1'\ISA S_1 \} 
% \end{array}
% \]
% where the $B_i$'s are $\dlliter$ basic concepts and the $R_i$'s are $\dlliter$ basic roles.

% \ 

% Then, given a $\dlliter$ TBox $\T$, a policy $\P$ of binary EDs, and a set of atoms $\Z$, we define the formula $\isdisclformula(\Z,\T,\P)$ as follows:
% \[
% \begin{array}{r@{}l}
% \displaystyle
% \conj(\Z)\,\land
% & \displaystyle
% \bigwedge_{\substack{
%     B_1\ISA B_2 \in \policyexp_c(\P,\T)\;\land\\
%     B_1(t)\in\conj(\Z)
% }}
% %\bigwedge_{\tau \in \policyexp(\P,\T)}
% %\bigwedge_{\sigma\in \subst(\conj(\Z),\body(\tau))}
% \perfectref(B_2(t),\T)
% \,\land \\
% & \displaystyle\;
% \bigwedge_{\substack{
%     S_1\ISA S_2 \in \policyexp_r(\P,\T)\;\land\\
%     S_1(t_1,t_2)\in\conj(\Z)
% }}
% %\bigwedge_{\tau \in \policyexp(\P,\T)}
% %\bigwedge_{\sigma\in \subst(\conj(\Z),\body(\tau))}
% \perfectref(S_2(t_1,t_2),\T)
% \end{array}
% \]

As said above, this kind of policy enabled us not to re-implement a rewriting function $\ucqrew$ of Algorithm~\ref{alg:policy-expand} for linear EDs.
Formally, when $\P$ is a binary policy, instead of $\ucqrew(\body(\tau),\Sigma(\P,\T))$, we computed $\ucqrew(\body(\tau),\allowbreak\DL(\P^+)\cup\T)$ by exploiting the tree-witness rewriter also for this purpose.

One may wonder whether this restriction is sufficient to get back FO-rewritability. %, without introducing the full condition. 
However, the kind of policy used in the proof of Theorem~\ref{thm:binary-iga-lb} immediately rules out this possibility.
%However, the following theorem, along with the fact that $\aczero\subsetneq\NL$, immediately rules out this possibility.
Hence, we restrict our attention to the case of \emph{full} binary EDs, which is FO-rewritable by Theorem~\ref{thm:iga-aczero}.\footnote{We recall that the restriction to a single atom in the head does not actually decrease the expressiveness of full EDs (and thus of full binary EDs).} 
% although syntactically restricted to heads with a single atoms, full binary EDs are actually able to express conjunctions of atoms in their heads

%%%%%%%%%%%%%%%%%%%%%%%%%%%%%%%%%%
%%%%%%%%%%%%%%%%%%%%%%%%%%%%%%%%%%

\subsection{Policy Definition}

We ran our experiments using an acyclic policy $\polAcy$ (consisting of 6 EDs) and a binary policy $\polBin$ (consisting of 11 EDs), namely (we removed the universally quantified variables for improving their readability): %\nb{se occupano troppo spazio potremmo limitarci a rimandare il lettore al repository}
\newcommand{\ED}[4]{%#1: 
    %\forall #2\,(
    K\,#3\rightarrow K\,#4%)
    }
\newcommand{\ComputerScience}{\predSymbol{ComputerScience}}
\newcommand{\ElectiveCourse}{\predSymbol{ElectiveCourse}}
\newcommand{\Employee}{\predSymbol{Employee}}
\newcommand{\FineArts}{\predSymbol{FineArts}}
\newcommand{\FullProfessor}{\predSymbol{FullProfessor}}
\newcommand{\hasAlumnus}{\predSymbol{hasAlumnus}}
\newcommand{\hasCollaborationWith}{\predSymbol{hasCollaborationWith}}
\newcommand{\hasCollegeDiscipline}{\predSymbol{hasCollegeDiscipline}}
\newcommand{\hasDoctoralDegreeFrom}{\predSymbol{hasDoctoralDegreeFrom}}
\newcommand{\hasMajor}{\predSymbol{hasMajor}}
\newcommand{\hasMasterDegreeFrom}{\predSymbol{hasMasterDegreeFrom}}
\newcommand{\hasSameHomeTownWith}{\predSymbol{hasSameHomeTownWith}}
\newcommand{\isAdvisedBy}{\predSymbol{isAdvisedBy}}
\newcommand{\isAffiliatedOrganizationOf}{\predSymbol{isAffiliatedOrganizationOf}}
\newcommand{\isVisitingProfessorOf}{\predSymbol{isVisitingProfessorOf}}
\newcommand{\knows}{\predSymbol{knows}}
\newcommand{\Person}{\predSymbol{Person}}
\newcommand{\Professor}{\predSymbol{Professor}}
\newcommand{\Student}{\predSymbol{Student}}
\newcommand{\takesCourse}{\predSymbol{takesCourse}}
\newcommand{\teachesCourse}{\predSymbol{teachesCourse}}
\newcommand{\Utwo}{\predSymbol{U2}}
\newcommand{\Woman}{\predSymbol{Woman}}
{\small
%%%%%%%%%%%%%%%%%%%%%%%
%%%%%%%%%%%%%%%%%%%%%%%
\[
\begin{array}{r@{\,}l}
    %\polAcy =
    \{
    & \ED{\tau_1^a}{y}{\exists x\,\isAdvisedBy(x,y)}{\Woman(y)},\\
    & \ED{\tau_2^a}{y}{\exists x\,(\hasAlumnus(x,y)\land\hasMasterDegreeFrom(y,x)}{\FullProfessor(y)},\\
    & \ED{\tau_3^a}{x,y}{(\takesCourse(x,y)\land\Student(x))}{\bot},\\
    & \ED{\tau_4^a}{x,y}{(\hasCollaborationWith(x,y)\land\Student(x))}{\bot},\\
    & \ED{\tau_5^a}{x,y,z}{\isAdvisedBy(x,y)\land\hasMasterDegreeFrom(x,z)}{\hasMajor(x, \ComputerScience)},\\
    & \ED{\tau_6^a}{x,y}{(\teachesCourse(x,y)\land\FullProfessor(x))}{\hasDoctoralDegreeFrom(x, \Utwo)}
    \,\} ,\\
    %%%%%%%%%%%%%%%%%%%%%%%
    %%%%%%%%%%%%%%%%%%%%%%%
    %\polBin =
    \{
    & \ED{\tau_1^b}{x,y}{\isAffiliatedOrganizationOf(x,y), \hasCollegeDiscipline(x,\FineArts)}{\bot},\\
    & \ED{\tau_2^b}{x}{\Professor(x)}{\Woman(x)},\\
    & \ED{\tau_3^b}{x}{\teachesCourse(x,y)}{\FullProfessor(x)},\\
    & \ED{\tau_4^b}{x}{\isVisitingProfessorOf(x,y)}{\bot},\\
    & \ED{\tau_5^b}{y}{\takesCourse(x,y)}{\ElectiveCourse(y)},\\
    & \ED{\tau_6^b}{x}{\Person(x)}{\Employee(x)},\\
    & \ED{\tau_7^b}{x,y}{\hasAlumnus(x,y)}{\hasMasterDegreeFrom(y,x)},\\
    & \ED{\tau_8^b}{x}{\hasCollaborationWith(x,y)}{\Professor(x)},\\
    & \ED{\tau_9^b}{x}{\hasSameHomeTownWith(x,y)}{\Employee(x)},\\
    & \ED{\tau_{10}^b}{x}{\knows(x,y)}{\Professor(x)},\\
    & \ED{\tau_{11}^b}{y}{\knows(x,y)}{\Professor(y)}
    \,\} .
\end{array}
\]
%%%%%%%%%%%%%%%%%%%%%%%
%%%%%%%%%%%%%%%%%%%%%%%
}
%%
%We also defined two reduced versions of $\polAcy$ and $\polBin$, respectively $\polAcyRed=\{\tau_4^a,\tau_5^a,\tau_6^a\}$ and $\polBinRed=\{\tau_6^b,\tau_7^b,\tau_8^b,\tau_9^b,\tau_{10}^b,\tau_{11}^b\}$, where every $\tau_i^j$ is the $i$-th dependency occurring in the policy $\P_j$.
We also defined $\polAcyRed$ as a reduced version of $\polAcy$, containing its last 3 EDs, and $\polBinRed$ as a reduced version of $\polBin$, containing its last 6 EDs.

%%%%%%%%%%%%%%%%%%%%%%%%%%%

\subsection{Results}
\label{sec:results}
The results of our experiments are reported in Table~\ref{tab:all-results}. The symbol $\P_\emptyset$ indicates the absence of a policy: this is the configuration that we used as a baseline.

The main indications provided by these results are the following:  
\begin{itemize}
    \item 
    In most cases, the evaluation time $t_e$ is acceptable (it remains on the order of seconds). The only critical query is the seventh one, which takes several minutes to execute. 
    
    \item 
    The time $t_r$ necessary for computing the rewritten query is always less than or equal to 3 seconds. As expected, for both \benchFive and \benchTen, such rewriting times are very close, as they do not depend on the ABox. 
    
    \item 
    For both acyclic and binary policies, the values of $t_r$ corresponding to smaller and larger policies are of comparable magnitude.
    
    \item 
    Observe that binary policies tend to remove more tuples than acyclic ones. This could be explained by the fact that EDs with fewer atoms in their body are more likely ``activated'' by the query.
\end{itemize}

% Disposizione verticale
\newcommand{\result}[3]{\makecell{#1\\#2\\#3}}

% Disposizione orizzontale
%\newcommand{\result}[2]{ #1 & #2 }

\newcommand{\queryrow}[1]{%
    %\multicolumn{1}{|c|}{\multirow[c]{2}{*}{$q_#1$}}
    \parbox[c][2\baselineskip][c]{0.4cm}{\hspace{1pt}$q_{#1}$} &
    \multicolumn{1}{|c|}{\makecell{$t_r$\\$t_e$\\\#}}
}

% Bordo verticale in grassetto leggero
\newcolumntype{;}{!{\vrule width 1pt}}

\begin{table}[H] % forcing the position
%\begin{table}[!htbp] % more flexible
\centering
\begin{tabular}{|cc|c;c|c;c|c||c;c|c;c|c|}
\hline
\multicolumn{2}{|c|}{\multirow{2}{*}{Query}}
& \multicolumn{5}{c||}{o2b\textsubscript{5}}
& \multicolumn{5}{c|}{o2b\textsubscript{10}}\\
\cline{3-12}
&
& $\P_\emptyset$ & $\polAcyRed$ & $\polAcy$ & $\polBinRed$ & $\polBin$
& $\P_\emptyset$ & $\polAcyRed$ & $\polAcy$ & $\polBinRed$ & $\polBin$
\\
\hline
\queryrow{1}
& \result{19}{513}{9228}
& \result{19}{526}{9228}
& \result{20}{547}{9228}
& \result{53}{683}{1367}
& \result{63}{958}{334}
& \result{15}{822}{19782}
& \result{20}{560}{19782}
& \result{28}{964}{19782}
& \result{51}{1394}{2948}
& \result{63}{1085}{730}
\\\hline
\queryrow{2}
& \result{50}{81}{18872}
& \result{249}{8688}{18736}
& \result{336}{7335}{14829}
& \result{476}{174}{5957}
& \result{511}{206}{5957}
& \result{65}{269}{44190}
& \result{243}{39426}{43889}
& \result{386}{51181}{33193}
& \result{873}{876}{13009}
& \result{553}{402}{13009}
\\\hline
\queryrow{3}
& \result{25}{4}{34}
& \result{41}{8}{34}
& \result{42}{6}{34}
& \result{40}{2}{34}
& \result{109}{4}{28}
& \result{38}{8}{75}
& \result{32}{5}{75}
& \result{43}{9}{75}
& \result{62}{5}{75}
& \result{112}{6}{64}
\\\hline
\queryrow{4}
& \result{21}{6}{0}
& \result{40}{7}{0}
& \result{35}{2}{0}
& \result{52}{4}{0}
& \result{133}{3}{0}
& \result{33}{7}{0}
& \result{33}{5}{0}
& \result{49}{7}{0}
& \result{68}{5}{0}
& \result{97}{3}{0}
\\\hline
\queryrow{5}
& \result{21}{17}{3574}
& \result{554}{30699}{2020}
& \result{473}{29679}{2020}
& \result{250}{1823}{952}
& \result{319}{1003}{264}
& \result{29}{44}{6564}
& \result{482}{110763}{3676}
& \result{917}{124305}{3676}
& \result{451}{5056}{1696}
& \result{334}{1283}{394}
\\\hline
\queryrow{6}
& \result{18}{59}{16236}
& \result{114}{60661}{15834}
& \result{115}{28564}{7811}
& \result{161}{81}{0}
& \result{116}{230}{0}
& \result{25}{235}{35889}
& \result{148}{283020}{35075}
& \result{109}{141173}{17481}
& \result{263}{333}{0}
& \result{193}{282}{0}
\\\hline
\queryrow{7}
& \result{75}{66}{5489}
& \result{2029}{198641}{5489}
& \result{1976}{187908}{5091}
& \result{1378}{403}{5489}
& \result{1847}{496}{3292}
& \result{88}{334}{11969}
& \result{2205}{993701}{11969}
& \result{2244}{1119538}{10971}
& \result{2222}{1343}{11969}
& \result{1801}{812}{7241}
\\\hline
\queryrow{8}
& \result{22}{62}{17904}
& \result{52}{75}{17904}
& \result{48}{63}{17904}
& \result{257}{615}{14668}
& \result{263}{647}{14668}
& \result{21}{186}{39278}
& \result{49}{112}{39278}
& \result{44}{143}{39278}
& \result{399}{1548}{32350}
& \result{272}{963}{32350}
\\\hline
\queryrow{9}
& \result{135}{31}{1698}
& \result{1711}{1412}{1539}
& \result{2302}{1313}{1539}
& \result{1765}{137}{0}
& \result{3430}{129}{0}
& \result{195}{93}{3434}
& \result{1778}{10230}{3196}
& \result{2290}{10944}{3196}
& \result{2877}{598}{0}
& \result{3144}{332}{0}
\\\hline
\queryrow{10}
& \result{22}{78}{642}
& \result{1243}{676}{122}
& \result{1220}{3}{0}
& \result{184}{110}{642}
& \result{337}{159}{144}
& \result{32}{297}{1413}
& \result{1351}{4753}{258}
& \result{1433}{3}{0}
& \result{380}{478}{1413}
& \result{477}{270}{335}
\\\hline
\end{tabular}
%\caption{All the evaluation results. For each entry, we reported the number of returned tuples (\#) and the time expressed in milliseconds (ms).}
\vspace*{1mm}
\caption{All the evaluation results. Each entry reports the rewriting time ($t_r$) and evaluation time ($t_e$) expressed in milliseconds, plus the number of returned tuples (\#).}
\label{tab:all-results}
\end{table}

    \section{Conclusions}
\label{sec:conclusions}

In this work, we investigated CQE under policies expressed via epistemic dependencies (EDs), focusing on the use of ground atom (GA) censors for safe information disclosure. We investigated IGA-entailment, a semantic relation based on the intersection of all optimal GA censors, and analyzed its data complexity in the presence of ED-based policies when the TBox is expressed in $\dlliter$.
Our results show that the intersection remains safe for full EDs, a subclass of particular interest. We established that IGA-entailment is not FO-rewritable in general, which led us to the definition of the subclass of full and expandable EDs for which FO-rewriting is feasible, and we introduced a rewriting algorithm that works for $\dlliter$ ontologies. We validated our approach through a prototype implementation evaluated using the OWL2Bench benchmark, showing its practical feasibility in diverse evaluation scenarios.

As for future work, we are currently working on the definition of practical algorithms for some of the non-FO-rewritable cases of IGA-entailment analyzed in the paper. Another interesting research direction is towards extending the FO-rewritable cases identified by our analysis: in particular, we would like to focus either on further subclasses of EDs, or on (subclasses of) policy languages that go beyond the expressiveness of EDs. In addition, it would be interesting to study the complexity of CQE under EDs for TBoxes expressed in DLs different from $\dlliter$.
Finally, it would be of practical importance to extend the present approach to the framework of ontology-based data access (OBDA).

    \iflong\else
        \input{supplemental-material-statement}
    \fi

    % \ifcameraready
    %     \input{acknowledgements}
    % \fi
    
    \bibliographystyle{splncs04}
    \bibliography{bibliography/strings,bibliography/bibliography,bibliography/w3c}

\begin{thebibliography}{10}
\providecommand{\url}[1]{\texttt{#1}}
\providecommand{\urlprefix}{URL }
\providecommand{\doi}[1]{https://doi.org/#1}

\bibitem{AbHV95}
Abiteboul, S., Hull, R., Vianu, V.: Foundations of Databases. Addison Wesley Publ.\ Co. (1995)

\bibitem{BCMNP07}
Baader, F., Calvanese, D., McGuinness, D., Nardi, D., Patel-Schneider, P.F. (eds.): The Description Logic Handbook: {T}heory, Implementation and Applications. Cambridge University Press, 2nd edn. (2007)

\bibitem{BLMS11}
Baget, J., Lecl{\`{e}}re, M., Mugnier, M., Salvat, E.: On rules with existential variables: Walking the decidability line. Artificial Intelligence  \textbf{175}(9-10),  1620--1654 (2011)

\bibitem{Bisk00}
Biskup, J.: For unknown secrecies refusal is better than lying. Data and Knowledge Engineering  \textbf{33}(1),  1--23 (2000)

\bibitem{BiBo04}
Biskup, J., Bonatti, P.A.: Controlled query evaluation for enforcing confidentiality in complete information systems. Int. J. Inf. Sec.  \textbf{3}(1),  14--27 (2004)

\bibitem{Bona22}
Bonatti, P.A.: A false sense of security. Artificial Intelligence  \textbf{310}(103741) (2022)

\bibitem{BoSa13}
Bonatti, P.A., Sauro, L.: A confidentiality model for ontologies. In: Proc.\ of the 12th Int.\ Semantic Web Conf.\ (ISWC). Lecture Notes in Computer Science, vol.~8218, pp. 17--32. Springer (2013)

\bibitem{CGL12}
Cal{\`{\i}}, A., Gottlob, G., Lukasiewicz, T.: A general {Datalog}-based framework for tractable query answering over ontologies. J.\ of Web Semantics  \textbf{14},  57--83 (2012)

\bibitem{CDLLR07b}
Calvanese, D., De~Giacomo, G., Lembo, D., Lenzerini, M., Rosati, R.: {EQL-Lite}: Effective first-order query processing in description logics. In: Proc.\ of the 20th Int.\ Joint Conf.\ on Artificial Intelligence (IJCAI). pp. 274--279. Morgan Kaufmann Publishers Inc. (2007)

\bibitem{CDLLR07}
Calvanese, D., De~Giacomo, G., Lembo, D., Lenzerini, M., Rosati, R.: Tractable reasoning and efficient query answering in description logics: The \textit{DL-Lite} family. J.\ of Automated Reasoning  \textbf{39}(3),  385--429 (2007)

\bibitem{CLMRS20}
Cima, G., Lembo, D., Marconi, L., Rosati, R., Savo, D.F.: Controlled query evaluation in ontology-based data access. In: The Semantic Web - {ISWC} 2020 - 19th International Semantic Web Conference, Athens, Greece, November 2-6, 2020, Proceedings, Part {I}. Lecture Notes in Computer Science, vol. 12506, pp. 128--146. Springer (2020)

\bibitem{CLMRS24}
Cima, G., Lembo, D., Marconi, L., Rosati, R., Savo, D.F.: Enhancing controlled query evaluation through epistemic policies. In: Proc.\ of the 33th Int.\ Joint Conf.\ on Artificial Intelligence (IJCAI). pp. 3307--3314. Int.\ Joint Conf.\ on Artificial Intelligence Organization (Aug 2024)

\bibitem{CLMRS24-SNCS}
Cima, G., Lembo, D., Marconi, L., Rosati, R., Savo, D.F.: A gentle introduction to controlled query evaluation in {DL-Lite} ontologies. Springer Nature Computer Science  \textbf{5}(4), ~335 (2024)

\bibitem{CLMRS25-JoWS}
Cima, G., Lembo, D., Marconi, L., Rosati, R., Savo, D.F.: Indistinguishability in controlled query evaluation over prioritized description logic ontologies. J.\ of Web Semantics  \textbf{84},  100841 (2025)

\bibitem{CLRS24}
Cima, G., Lembo, D., Rosati, R., Savo, D.F.: Controlled query evaluation in description logics through consistent query answering. Artificial Intelligence  \textbf{334},  104176 (2024)

\bibitem{CL20}
Console, M., Lenzerini, M.: Epistemic integrity constraints for ontology-based data management. In: Proc.\ of the 34th AAAI Conf.\ on Artificial Intelligence (AAAI). pp. 2790--2797. {AAAI} Press (2020)

\bibitem{CHMP*08}
Cuenca~Grau, B., Horrocks, I., Motik, B., Parsia, B., Patel-Schneider, P., Sattler, U.: {OWL~2}: {T}he next step for {OWL}. J.\ of Web Semantics  \textbf{6}(4),  309--322 (2008)

\bibitem{CKKZ13}
Cuenca~Grau, B., Kharlamov, E., Kostylev, E.V., Zheleznyakov, D.: Controlled query evaluation over {OWL} 2 {RL} ontologies. In: Proc.\ of the 12th Int.\ Semantic Web Conf.\ (ISWC). pp. 49--65 (2013)

\bibitem{KKZ12}
Kikot, S., Kontchakov, R., Zakharyaschev, M.: Conjunctive query answering with {OWL~2~QL}. In: Proc.\ of the 13th Int.\ Conf.\ on the Principles of Knowledge Representation and Reasoning (KR). {AAAI} Press (2012)

\bibitem{KLMT15}
K{\"{o}}nig, M., Lecl{\`{e}}re, M., Mugnier, M., Thomazo, M.: Sound, complete and minimal {UCQ}-rewriting for existential rules. Semantic Web J.  \textbf{6}(5),  451--475 (2015)

\bibitem{M98}
van Melkebeek, D.: Deterministic and randomized bounded truth-table reductions of {P}, {NL}, and {L} to sparse sets. J. Comput. Syst. Sci.  \textbf{57}(2),  213--232 (1998)

\bibitem{W3Crec-OWL2-Profiles}
Motik, B., Cuenca~Grau, B., Horrocks, I., Wu, Z., Fokoue, A., Lutz, C.: {OWL~2} {W}eb {O}ntology {L}anguage profiles (second edition). {W3C} {R}ecommendation, World Wide Web Consortium (Dec 2012), available at \protect\url{http://www.w3.org/TR/owl2-profiles/}

\bibitem{Papa94}
Papadimitriou, C.H.: Computational Complexity. Addison Wesley Publ.\ Co. (1994)

\bibitem{MKZ13}
Rodriguez{-}Muro, M., Kontchakov, R., Zakharyaschev, M.: Ontology-based data access: Ontop of databases. In: Proc.\ of the 12th Int.\ Semantic Web Conf.\ (ISWC). Lecture Notes in Computer Science, vol.~8218, pp. 558--573. Springer (2013)

\bibitem{SBM20}
Singh, G., Bhatia, S., Mutharaju, R.: {OWL2Bench}: {A} benchmark for {OWL}~2 reasoners. In: Proc.\ of the 19th Int.\ Semantic Web Conf.\ (ISWC). Lecture Notes in Computer Science, vol. 12507, pp. 81--96. Springer (2020)

\bibitem{owl2bench-queries}
Singh, G., Bhatia, S., Mutharaju, R.: {OWL2Bench} {SPARQL} queries (May 2020), \url{https://zenodo.org/records/3838735}

\end{thebibliography}

    % \newpage
    % \input{appendix}
    
\end{document}